\documentclass[preprint]{elsarticle}

\usepackage{times}
\usepackage{epsfig}
\usepackage{graphicx}
\usepackage{amsmath}
\usepackage{amssymb}
\usepackage{booktabs}
\usepackage{multirow}
\usepackage{subcaption}
\usepackage[export]{adjustbox}
\usepackage{float}
\usepackage{dashrule}
\usepackage{arydshln}
\usepackage[utf8x]{inputenc}
\usepackage[space]{grffile}

\newcommand{\etal}{\textit{et al.}}

\begin{document}

\begin{frontmatter}

\title{GANILLA: Generative Adversarial Networks for Image to Illustration Translation}

\author[hacet]{Samet Hicsonmez\corref{cor1}}
\ead{samethicsonmez@hacettepe.edu.tr}
\cortext[cor1]{Corresponding Author}
\author[odtu]{Nermin Samet}
\ead{nermin.samet@metu.edu.tr}
\author[odtu]{Emre Akbas}
\ead{emre@ceng.metu.edu.tr}
\author[hacet]{Pinar Duygulu}
\ead{pinar@cs.hacettepe.edu.tr}

\address[hacet]{Hacettepe University, Ankara, Turkey}
\address[odtu]{Middle East Technical University, Ankara, Turkey}


\begin{abstract}

In this paper, we explore illustrations in children’s books as a new domain in unpaired image-to-image translation. We show that although the current state-of-the-art image-to-image translation models successfully transfer either the style or the content, they fail to transfer both at the same time. We propose a new generator network to address this issue and show that the resulting network strikes a better balance between style and content.

There are no well-defined or agreed-upon evaluation metrics for unpaired image-to-image translation. So far, the success of image translation models has been based on subjective, qualitative visual comparison on a limited number of images. To address this problem, we propose a new framework for the quantitative evaluation of image-to-illustration models, where both content and style are taken into account using separate classifiers. In this new evaluation framework, our proposed model performs better than the current state-of-the-art models on the illustrations dataset. Our code and pretrained models can be found at https://github.com/giddyyupp/ganilla.

\end{abstract}


\end{frontmatter}


\section{Introduction}

Image-to-image style transfer has received increasing attention since Gatys et al.’s pioneering work~\cite{gatsy}. Researchers have developed various approaches for the problem, including paired~\cite{gatsy, gatsy2} and unpaired~\cite{cyclegan, dualgan, cartoongan} transfer, optimization based online methods~\cite{gatsy, onio1, onio2, onio3} and offline methods based on convolutional neural networks (CNN)~\cite{ofio1, ofio2, ofio3, ofio1-4} or generative adversarial networks (GAN)~\cite{cyclegan, dualgan, cartoongan, stargan, levent, pix2pix}. 
Style transfer has been applied to various domains including to transform natural images to art paintings~\cite{lee2018diverse, cyclegan, cho2019image, amodio2019travelgan} and reverse~\cite{tomei2019art2real, amodio2019travelgan}, to transfer the style of a certain animal to another animal~\cite{cyclegan, liu2019few, liu2017unsupervised}, or to transform certain properties (e.g. weather, season of the year) of the scene~\cite{levent, lee2018diverse, liu2017unsupervised, huang2018multimodal}, or to transfer aerial images to maps~\cite{pix2pix}, or to transfer~\cite{sketchygan, alharbi2019latent} or complete~\cite{liu2019sketchgan, ghosh2019isketchnfill} sketches or to convert face images to sketches~\cite{dualgan} and for cartoonization~\cite{cartoongan}.

\begin{figure}[h]
\centering
\begin{subfigure}[b]{0.18\textwidth}
   \caption{Input Image}
\includegraphics[width=1.0\textwidth]{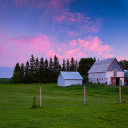}
\vspace{0.05 cm}
\includegraphics[width=1.0\textwidth]{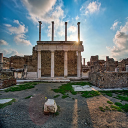}
\end{subfigure}
\begin{subfigure}[b]{0.18\textwidth}
   \caption{CycleGAN}
\includegraphics[width=1.0\textwidth]{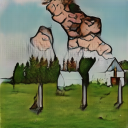}
\vspace{0.05 cm}
\includegraphics[width=1.0\textwidth]{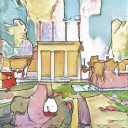}
\end{subfigure}
\begin{subfigure}[b]{0.18\textwidth}
   \caption{DualGAN}
\includegraphics[width=1.0\textwidth]{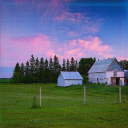}
\vspace{0.05 cm}
\includegraphics[width=1.0\textwidth]{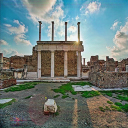}
\end{subfigure}
\begin{subfigure}[b]{0.18\textwidth}
   \caption{GANILLA}
\includegraphics[width=1.0\textwidth]{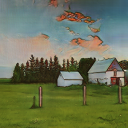}
\vspace{0.05 cm}
\includegraphics[width=1.0\textwidth]{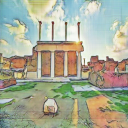}
\end{subfigure}
\caption{Example outputs from CycleGAN~\cite{cyclegan}, DualGAN~\cite{dualgan} and our method (GANILLA) using different styles. CycleGAN and GANILLA generate images in illustrator’s style, but DualGAN fails to transfer style. However, CycleGAN fails to preserve content of the input image e.g. it randomly places animal faces in the air. Our method preserves content as well while transferring the style.}
\label{fig:initial-res}
\end{figure}

This paper aims to introduce a new domain to the style transfer literature: illustrations in children’s books (just “illustrations” for short from here on) (Fig.~\ref{fig:sample-ill}). We claim that this is a new domain because illustrations are qualitatively different than art paintings and cartoons. Illustrations do contain objects (e.g. mountains, people, toys, trees, etc.), yet the abstraction level might be much higher than that of paintings and cartoons. Existing models fall short in dealing with the complex balance between abstraction style and content of the illustrations. For example, despite the impressive results of CycleGAN in other domains~\cite{cyclegan, pix2pixHD}, for illustrations we observed that the success at transferring the style of the artist is not reflected in transferring the content, due to the high level abstractions of the objects. On the other hand, most of the time DualGAN~\cite{dualgan} is successful at preserving the content of the image, but falls short transferring the style   (see Figure~\ref{fig:initial-res}). As empirically confirmed with our experiments, other state-of-the-art unpaired style transfer methods also fail to provide satisfactory results for illustrations.

Our goal is to develop a model that can generate appealing yet content preserving illustrations from a given natural image by transferring the style of a given illustration artist. To this end, we take the “unpaired” approach where two unaligned separate sets, one for the source domain (natural images) and one for the target (illustrations) are needed. We built upon our existing illustrations dataset~\cite{draw} that was used to classify illustrators and almost doubled its size to create the most extensive illustrations dataset. Our dataset contains 9448 illustrations coming from 363 different books and 24 different artists.

To address the issue of balancing style versus content, we propose two changes to the current state-of-the-art models. First, we propose a new generator network that down samples the feature map at each residual layer. Second, to better transfer the content, we propose to merge low level features with high level features by using skip connections and upsampling. Low level features generally contain edge like information which helps the generated image to preserve the structure of the input image. Our ablation experiments confirmed the effectiveness of these proposals.

One major problem with the unpaired style transfer approach is the evaluation part. Typically, the success of image-to-image translation models are evaluated qualitatively using a limited number of images or via user studies. Although there are some quantitative metrics~\cite{gan-metrics} for other image generation tasks, it is not possible to use them in this domain directly since there is no paired ground-truth for generated images. We argue that the evaluation must consider both the style and content aspects simultaneously. To this end, we propose a quantitative evaluation framework that is based on content and style classifiers and experimentally show that it produces reasonable results.

In summary, our paper makes contributions at three different levels of the style transfer problem: dataset, architecture and evaluation measures. Specifically, 
\begin{itemize}
\setlength{\itemsep}{0pt plus 0pt}
\item We explore illustrations in children’s books as a new domain in the image-to-image style and content transfer.  
\item We present the most extensive illustrations dataset with almost 9500 illustrations of 24 artists.  
\item We propose a novel generator network which strikes a good balance between style and content.  
\item We propose a new framework to quantitatively evaluate image generation models in terms of content and style. Using this new evaluation measure, our proposed model performs better than current state-of-the-art models on the most extensive illustrations dataset available. 
\end{itemize}

\section{Related Work}

It is possible to divide the current best practices for image style transfer into two. First approach, Neural Style Transfer (NST), uses CNNs to stylize an input image with the given style image which could be a painting or an illustration. The second approach utilizes GANs to synthesize stylized images. There are very comprehensive survey papers for both CNN~\cite{survey1} and GAN~\cite{survey2} based methods. Here we summarize the GAN based unpaired image-to-image translation methods which are the most relevant group of work to our model.

GANs~\cite{gan, gan2} are extensively used to generate images where generated image could be a handwritten digit~\cite{gan}, a bedroom image~\cite{dcgan} or be conditioned with a text~\cite{levent}. In the image-to-image translation domain, very successful methods have been proposed~\cite{pix2pix, cartoongan, cyclegan, dualgan, discogan, stargan}. There are two major groups of methods in this category: paired and unpaired.

The “paired” group takes the conventional supervised learning approach, where explicit input-target pairs are needed. A prominent example in the paired group is the Pix2pix~\cite{pix2pix} model which uses different image pairs for various tasks such as ``semantic labels to street scene” or ``day to night.” Collecting such datasets requires too much effort. To overcome this disadvantage, the second group requires unpaired image sets~\cite{cyclegan, dualgan, cartoongan, stargan}. These methods use two separate image collections, i.e. one for the source domain and one for the target domain, without explicitly pairing any two images.

CycleGAN~\cite{cyclegan} and DualGAN~\cite{dualgan} are pioneering works in the unpaired image-to-image translation approach. They both utilize a cyclic framework which consists of a couple of generators and discriminators. First couple learns a mapping from the source to target, while the second one learns a reverse mapping. In CycleGAN, they successfully tackle various tasks such as converting natural images to paintings, apples to oranges, and horses to zebras. 
In DualGAN, they successfully convert sketches to faces and day photos to photos at night.

On the other hand, CartoonGAN~\cite{cartoongan} presents a new training framework to replace the cyclic structure of CycleGAN and introduces new loss functions. In their framework, content information is extracted from different high level feature maps of the VGG network. They used these feature maps to feed their loss function called the semantic content loss. As the target image set, they use edge smoothed fake cartoon images and unprocessed cartoon images as well. These images are very different in terms of both content and style from the illustrations in our dataset.

\section{GANILLA}

In our preliminary experiments on image-to-illustration problem, we observed that the current unpaired image-to-image translation models~\cite{cyclegan, dualgan, cartoongan} fail to transfer style and content at the same time. We designed a new generator network in such a way that it preserves the content and transfers the style at the same time. In the following, we first describe our model in detail and then compare its structure with three state-of-the-art models. We also present the two ablation models which guided our design.

\subsection{Details of GANILLA}

\begin{figure*}[h]
\centering
\includegraphics[width=\textwidth]{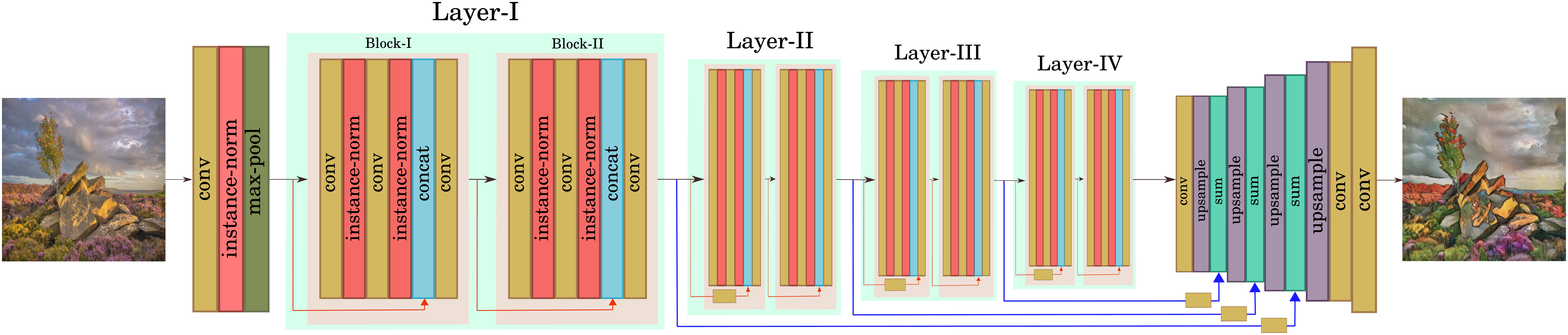}
\caption{GANILLA generator network. Our model uses concatenative skip connections for downsampling (from image up to Layer-IV). Then, the output of Layer-IV is sequentially upsampled while lower-level features from the downsampling stage are added to it via long, skip connections (blue arrows). The final output is a 3-channel stylized image. }
\label{fig:ganilla-model}
\end{figure*}

\begin{figure*}[h]
\centering
\includegraphics[width=\textwidth]{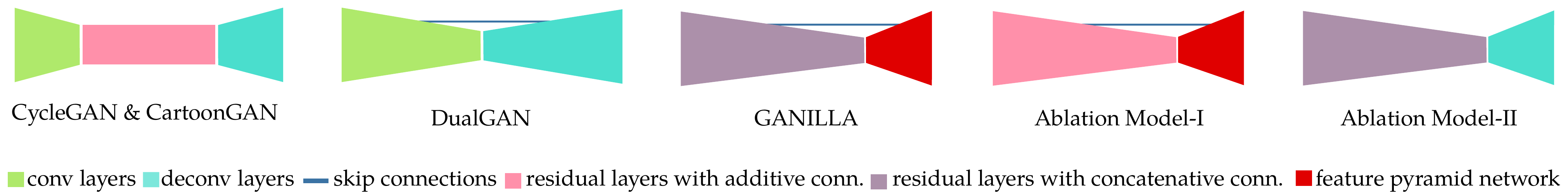}
\caption{High-level architectural description of state-of-the-art models (CycleGAN, CartoonGAN, DualGAN), our model (GANILLA) and the two ablation models we experimented with.}
\label{fig:model-sum}
\end{figure*}

A high-level architectural description of our generator network, GANILLA, is presented in Figure~\ref{fig:model-sum} along with the current state-of-the-art models and our two ablation models. GANILLA utilizes low-level features to preserve content while transferring the style. Our model consists of two stages (Figure~\ref{fig:ganilla-model}): the downsampling stage and the upsampling stage. The downsampling stage is a modified ResNet-18~\cite{resnet} network with the following modifications: In order to integrate low-level features at the downsampling stage, we concatenate features from previous layers at each layer of downsampling. Since low-level layers incorporate information like morphological features, edge and shape, they ensure that transferred image has the substructure of the input content.

More specifically, the downsampling stage starts with one convolution layer having $7 \times 7$ kernel followed by instance norm~\cite{IN}, ReLU and max pooling layers. Then continues with four layers where each layer contains two residual blocks. Each residual block starts with one convolution layer followed by instance norm and ReLU layers. Then, one convolution and an instance normalization layer follow. We concatenate the output with residual block input. Finally, we feed this concatenated tensor to the final convolution and ReLU layers. We halve feature map size in each layer except Layer-I using convolutions with stride of $2$. All convolution layers inside the residual layers have $3 \times 3$ kernels.

In the upsampling stage, we feed lower-level features using the outputs of each layer in downsampling stage through long, skip connections (blue arrows in Figure~\ref{fig:ganilla-model}) to the summation layers before the upsampling (Nearest Neighbor) operations. These connections help preserve the content.
In detail, upsampling stage contains four consecutive convolution, upsample and summation layers. First, output of Layer-IV is fed through convolution and upsample layers to increase feature map size in order to match feature size of previous layer. Convolution and upsample operations are conducted on consecutive layer’s outputs. All convolution filters in upsample stage have $1 \times 1$ kernels. Finally, one convolution layer with $7 \times 7$ kernel is used to output 3 channel translated image.

Our discriminator network is a $70 \times 70$ PatchGAN which is used for successful image to image translation models~\cite{cyclegan,pix2pix,Ledig_2017,li2016precomputed}. It consists of three blocks of convolutions where each block contains two convolution layers. We start with filter size of 64 for the first block, and double it for each consecutive block.

We follow the idea of cycle-consistency~\cite{cyclegan, dualgan} to train our GANILLA model. Specifically, there are two couples of generator-discriminator models. The first set (\textit{G}) tries to map source images to target domain, while the second set (\textit{F}) takes input as the target domain images and tries to generate source images in a cyclic fashion.

Our loss function consists of two Minimax losses~\cite{goodfellow2014generative} for each Generator and Discriminator pair, and one cycle consistency loss~\cite{cyclegan, dualgan}. Cycle consistency loss tries to ensure that a generated sample could be mapped back to source domain. We use \textit{${L}_{1}$} distance for cycle consistency loss.   
When we give source domain images to generator \textit{F} , we expect no change on them since they already correspond to source domain. A similar situation applies when we feed the generator \textit{G} with target domain images. This technique is first introduced by Taigman~\etal~\cite{taigman2016unsupervised} and used by CycleGAN~\cite{cyclegan} for image to painting translation tasks. We also use this identity paradigm (identity loss) with \textit{${L}_{1}$} distance function in our experiments. 
Our full objective function is to minimize the sum of these four loss functions.
For the details of the loss functions refer to~\cite{cyclegan}.

We used PyTorch~\cite{pytorch} to implement our models. As in other unpaired image-to-image transfer settings, GANILLA does not need paired images but two different image datasets, one for source and the other for target. For this dataset we use natural images as source domain, and illustration images as target domain. The details of the datasets will be provided in Section~\ref{sec:ill_dataset}. All train images (i.e. natural images and illustrations) are resized to $256\times256$ pixels. We train our models for 200 epoch using Adam solver with a learning rate of 0.0002. All networks were trained from scratch (i.e. not initialized with Imagenet weights). We conducted all our experiments on Nvidia Tesla V100 GPUs.

\begin{table}[h]
\centering
\caption{ Illustration dataset used in experiments.}
\begin{adjustbox}{max width=0.98\columnwidth}
\begin{tabular}{lccclccc}
\hline
Id & Illustrator & Book Cnt & Image Cnt & Id & Illustrator & Book Cnt & Image Cnt\\
\hline
AS        & Axel Scheffler         & 15         & 571         & PP        & Patricia Polacco         & 20 & 756        \\
DM         & David McKee         & 20         & 415         & RC        & Rosa Curto          & 10         & 336        \\        
KH         & Kevin Henkes         & 11         & 278          & SC         &Stephen Cartwright         & 25        & 405 \\
KP        & Korky Paul                  & 13         & 362         & SD        & Serap Deliorman        & 5         & 158         \\
MB          & Marc Brown         & 18         & 461        & TR         & Tony Ross                  & 21         & 521          \\
\hline
\end{tabular}
\end{adjustbox}
\label{tab:ill_data_set_info}
\end{table}
\subsection{Comparison with other models}
In Figure~\ref{fig:model-sum}, we summarized current state-of-the-art image to image single target translation GAN generator networks, our model and ablation models at high level architectural description. As it could be seen from the figure as a common approach convolutional layers for downsampling and deconvolutional layers for upsampling are used. In addition to those downsampling and upsampling layers in DualGAN, CycleGAN and CartoonGAN have additional flat residual layers with additive connections between downsampling and upsampling layers. 
In our generative model, we use residual layers with concatenative connections, and upsampling operations instead of deconvolutional layers. In our ablation models, we aimed to understand the effect of low-level features for both downsampling and upsampling parts. Our first ablation model is designed to observe how important the low-level features are  for the downsampling part. For this purpose, we replaced the concatenative connections with additive connections. We designed a second ablation model to test the effectiveness of low-level features for the upsampling part. In this ablation study, we replaced GANILLA’s upsampling layers with the block of upsampling layers of CycleGAN.

\begin{figure*}[h]
\captionsetup[subfigure]{labelformat=empty}
\centering
\setlength\tabcolsep{1.5pt} 
\begin{tabular}{rccrcc}
{AS} \includegraphics[width=0.14\textwidth,  ,valign=m, keepaspectratio,] {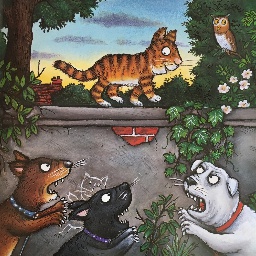} &
\includegraphics[width=0.14\textwidth,  ,valign=m, keepaspectratio,] {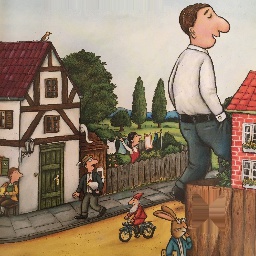} &
\includegraphics[width=0.14\textwidth,  ,valign=m,keepaspectratio,] {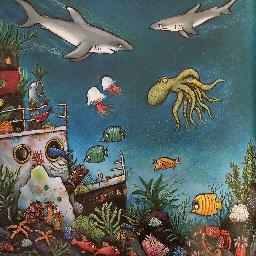} & 
{PP} \includegraphics[width=0.14\textwidth,  ,valign=m, keepaspectratio,] {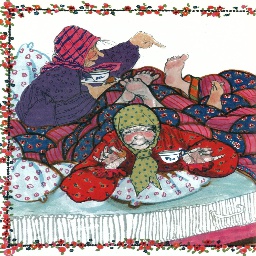} & 
\includegraphics[width=0.14\textwidth,  ,valign=m, keepaspectratio,] {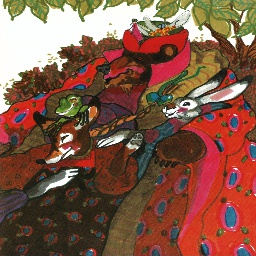} &
 \includegraphics[width=0.14\textwidth,  ,valign=m, keepaspectratio,] {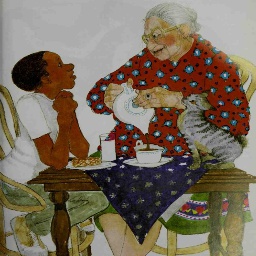}         \\

{DM}\includegraphics[width=0.14\textwidth,  ,valign=m, keepaspectratio,] {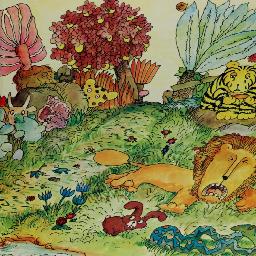} &
\includegraphics[width=0.14\textwidth,  ,valign=m, keepaspectratio,] {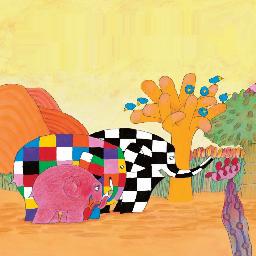} &
\includegraphics[width=0.14\textwidth,  ,valign=m,keepaspectratio,] {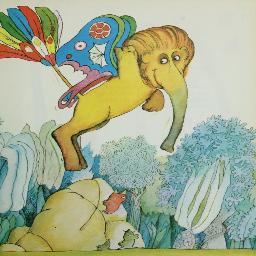} &
{RC} \includegraphics[width=0.14\textwidth,  ,valign=m, keepaspectratio,] {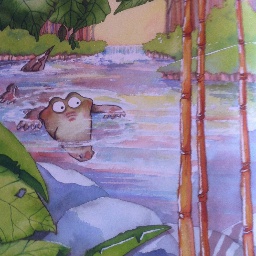} &
\includegraphics[width=0.14\textwidth,  ,valign=m, keepaspectratio,] {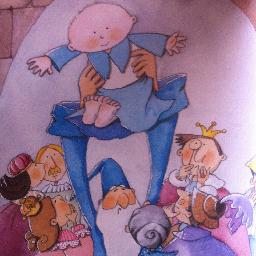} &
\includegraphics[width=0.14\textwidth,  ,valign=m, keepaspectratio,] {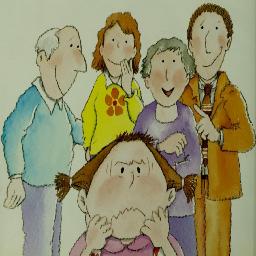} \\

{KH} \includegraphics[width=0.14\textwidth,  ,valign=m, keepaspectratio,] {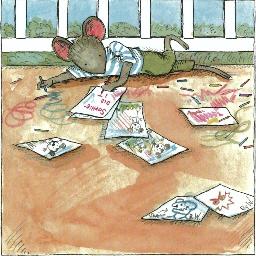} &
\includegraphics[width=0.14\textwidth,  ,valign=m, keepaspectratio,] {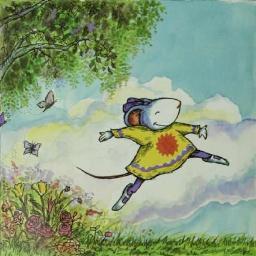} &
\includegraphics[width=0.14\textwidth,  ,valign=m, keepaspectratio,] {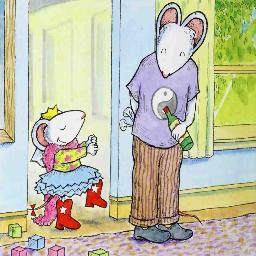} &
{SC} \includegraphics[width=0.14\textwidth,  ,valign=m, keepaspectratio,] {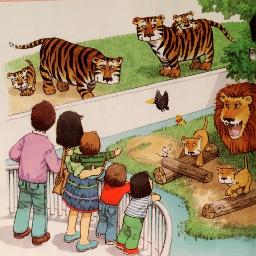} &
\includegraphics[width=0.14\textwidth,  ,valign=m, keepaspectratio,] {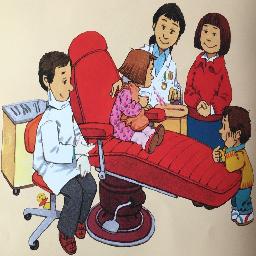} &
\includegraphics[width=0.14\textwidth,  ,valign=m, keepaspectratio,] {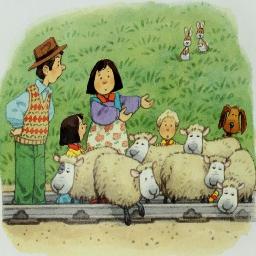} \\

{KP} \includegraphics[width=0.14\textwidth,  ,valign=m, keepaspectratio,] {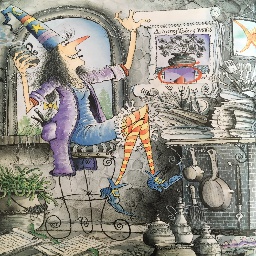} &
\includegraphics[width=0.14\textwidth,  ,valign=m, keepaspectratio,] {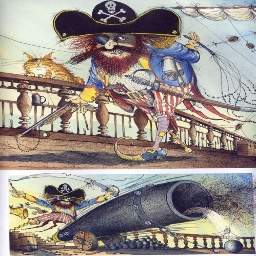} &
\includegraphics[width=0.14\textwidth,  ,valign=m, keepaspectratio,] {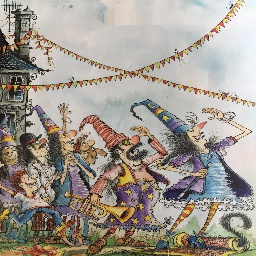} &
{SD} \includegraphics[width=0.14\textwidth,  ,valign=m, keepaspectratio,] {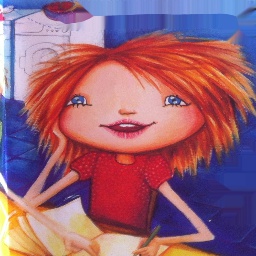} &
\includegraphics[width=0.14\textwidth,  ,valign=m, keepaspectratio,] {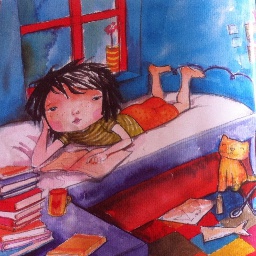} &
\includegraphics[width=0.14\textwidth,  ,valign=m, keepaspectratio,] {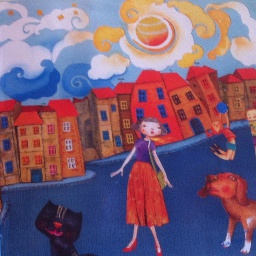} \\

{MB} \includegraphics[width=0.14\textwidth,  ,valign=m, keepaspectratio,] {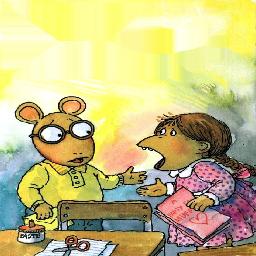} &
\includegraphics[width=0.14\textwidth,  ,valign=m, keepaspectratio,] {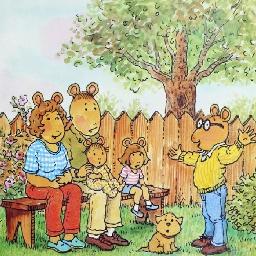} &
\includegraphics[width=0.14\textwidth,  ,valign=m, keepaspectratio,] {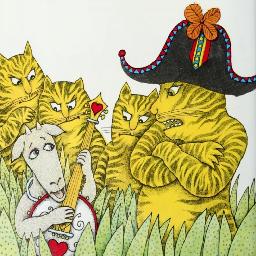} &
{TR} \includegraphics[width=0.14\textwidth,  ,valign=m, keepaspectratio,] {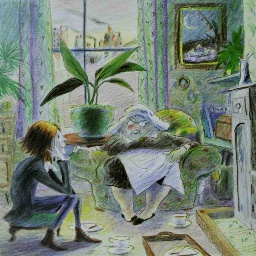} &
\includegraphics[width=0.14\textwidth,  ,valign=m, keepaspectratio,] {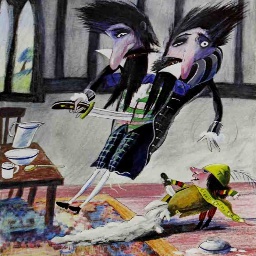} &
\includegraphics[width=0.14\textwidth,  ,valign=m, keepaspectratio,] {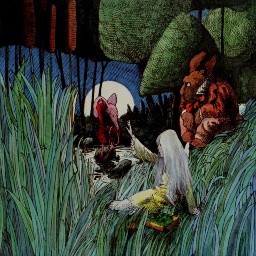} \\
\end{tabular}
\caption{Examples from illustration dataset. Note that, images are from different books with different characters. Generally, each illustrator has a distinct style which does not change much. However, as mostly observed with AS and RC on the third columns, sometimes it is very difficult even for a human to capture the style except for some fine details such as eyes. }
\label{fig:sample-ill}
\end{figure*}
\section{Datasets}
\label{sec:ill_dataset}
For training, we use 5402 natural images from CycleGAN training dataset~\cite{cyclegan} as the source domain, and we presented a new illustration dataset as the target domain. In testing, we use 751 images from CycleGAN test set.

For the illustration dataset, we extended the dataset in~\cite{draw} with new images almost doubling the original size. This extended dataset contains almost 9500 illustrations coming from 363 different books and 24 different artists. In order to train GAN models better, we increased the number of images for almost all illustrators. We collected new images by scraping the web and scanning books from public libraries. Our illustration dataset could be reproduced by scraping web based open libraries.
In this study, we use a subset of this dataset consisting illustrations from 10 artists who draw full page and complex scenes (see Table~\ref{tab:ill_data_set_info} for numbers and Figure~\ref{fig:sample-ill} for sample illustrations). We use the remaining 14 illustrators in our proposed quantitative evaluation framework in Section~\ref{sec:quantitative}. From now on, illustrators will be referred with their initials.

Our code, pretrained models and the scripts that reproduce the dataset can be found at https://github.com/giddyyupp/ganilla.

\section{Evaluation}

We compared our method with three state-of-the-art GAN methods that use unpaired data: CartoonGAN~\cite{cartoongan}, CycleGAN~\cite{cyclegan} and DualGAN~\cite{dualgan}. We used their official implementations that are publicly available. In the following, first we provide qualitative results and results of a user study (see Section~\ref{sec:qualitative}). The main difficulty in comparing the GAN methods is the lack of quantitative evaluations. In this study, we present new measures to handle this issue. 
Basically, there are two main factors which determine the quality of the GAN generated illustrations; 1) having target style and 2) preserving the content. We propose a new framework to evaluate GANs quantitatively. We introduce two CNNs, one for style and one for content. \textit{Style-CNN} aims to measure how well the results are in terms of style transfer. On the other hand, \textit{Content-CNN} aims to detect whether input content is preserved or not. Details of these two networks will be described in Section~\ref{sec:quantitative}.

In Table~\ref{tab:param_comparison} we compare GANILLA with other methods in terms of number of parameters and train time. We run CartoonGAN with a batch size of 4, and the other three methods with a batch size of 1. All models are trained for 200 epochs.

\begin{table}
\begin{center}
 \caption{Comparison with other methods in terms of number of parameters in their Generator networks and train time for one epoch on the illustrations dataset.}
 \label{tab:param_comparison}
 \resizebox{0.7\columnwidth}{!}{
 \begin{tabular}{lcccc}
   \toprule 
   & CartoonGAN & CycleGAN & DualGAN & GANILLA \\
  \cmidrule(lr){2-5}
   No of Params (Mil) & 11.1 & 11.4 & 54.1 & 7.2  \\
   Train Time (sec)   & 1400 & 1347 & 710 & 887 \\
   \bottomrule 
 \end{tabular}}
\end{center}
\end{table}

\begin{figure*}
\captionsetup[subfigure]{labelformat=empty}
\centering
\setlength\tabcolsep{1.5pt} 
\begin{tabular}{cccccc}
{Input} & {AS / PP}  & {DM / RC} & {KH / SC} & {KP / SD} & {MB / TR} \\
\multirow{2}{*}{\includegraphics[width=0.15\textwidth]{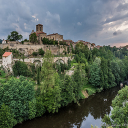}} &
\includegraphics[width=0.15\textwidth]{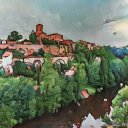} &
\includegraphics[width=0.15\textwidth]{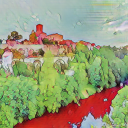} &
\includegraphics[width=0.15\textwidth]{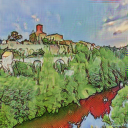} &
\includegraphics[width=0.15\textwidth]{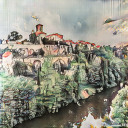}&
\includegraphics[width=0.15\textwidth]{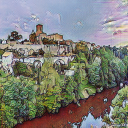} \\
&\includegraphics[width=0.15\textwidth]{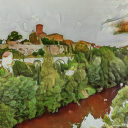} 
&\includegraphics[width=0.15\textwidth]{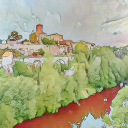} 
&\includegraphics[width=0.15\textwidth]{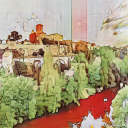} 
&\includegraphics[width=0.15\textwidth]{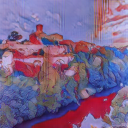} 
&\includegraphics[width=0.15\textwidth]{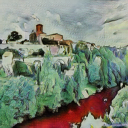} \\

\multirow{2}{*}{    \includegraphics[width=0.15\textwidth]{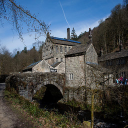}}&
\includegraphics[width=0.15\textwidth]{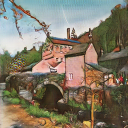} &
\includegraphics[width=0.15\textwidth]{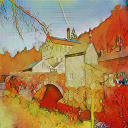} &
\includegraphics[width=0.15\textwidth]{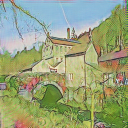} &
\includegraphics[width=0.15\textwidth]{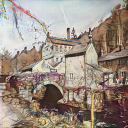}&
\includegraphics[width=0.15\textwidth]{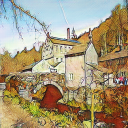} \\
&\includegraphics[width=0.15\textwidth]{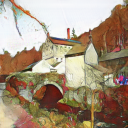} 
&\includegraphics[width=0.15\textwidth]{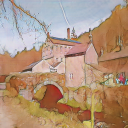} 
&\includegraphics[width=0.15\textwidth]{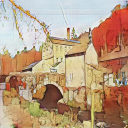} 
&\includegraphics[width=0.15\textwidth]{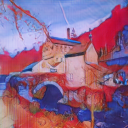} 
&\includegraphics[width=0.15\textwidth]{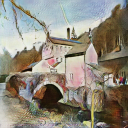} \\

\multirow{2}{*}{\includegraphics[width=0.15\textwidth]{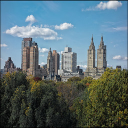}}&
\includegraphics[width=0.15\textwidth]{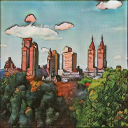} &
\includegraphics[width=0.15\textwidth]{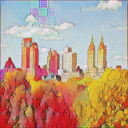} &
\includegraphics[width=0.15\textwidth]{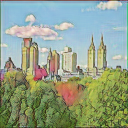} &
\includegraphics[width=0.15\textwidth]{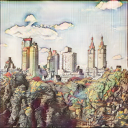}&
\includegraphics[width=0.15\textwidth]{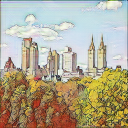} \\
&\includegraphics[width=0.15\textwidth]{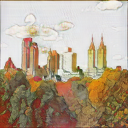} 
&\includegraphics[width=0.15\textwidth]{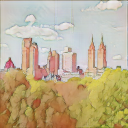} 
&\includegraphics[width=0.15\textwidth]{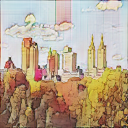} 
&\includegraphics[width=0.15\textwidth]{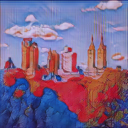} 
&\includegraphics[width=0.15\textwidth]{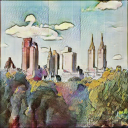} \\

\multirow{2}{*}{\includegraphics[width=0.15\textwidth]{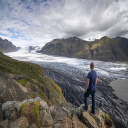}}&
\includegraphics[width=0.15\textwidth]{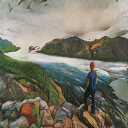} &
\includegraphics[width=0.15\textwidth]{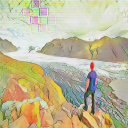} &
\includegraphics[width=0.15\textwidth]{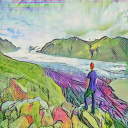} &
\includegraphics[width=0.15\textwidth]{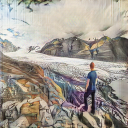} &
\includegraphics[width=0.15\textwidth]{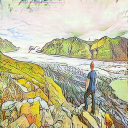} \\
&\includegraphics[width=0.15\textwidth]{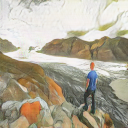} 
&\includegraphics[width=0.15\textwidth]{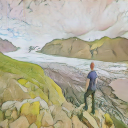} 
&\includegraphics[width=0.15\textwidth]{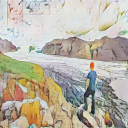} 
&\includegraphics[width=0.15\textwidth]{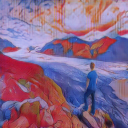} 
&\includegraphics[width=0.15\textwidth]{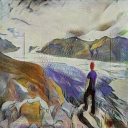} \\
\end{tabular}
\caption{Sample results of GANILLA. For a given test image, we present the style transfer results of all 10 illustrators in consecutive rows. Each column labeled as first and second row’s style id.}
\label{fig:our-results}
\end{figure*}

\begin{figure*}
\captionsetup[subfigure]{labelformat=empty}
\centering
\setlength\tabcolsep{1.5pt} 
\begin{tabular}{rcccc}
{Input}  & CartoonGAN & CycleGAN & DualGAN & GANILLA \\
{AS}  \includegraphics[width=0.14\textwidth,  ,valign=m, keepaspectratio,]{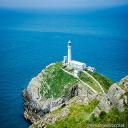} & 
\includegraphics[width=0.14\textwidth,  ,valign=m, keepaspectratio,]{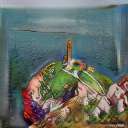}&
\includegraphics[width=0.14\textwidth,  ,valign=m, keepaspectratio,]{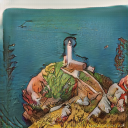}&
\includegraphics[width=0.14\textwidth,   ,valign=m, keepaspectratio,]{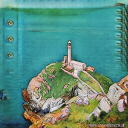}&
\includegraphics[width=0.14\textwidth,   ,valign=m, keepaspectratio,]{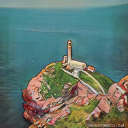}\\

{DM}  \includegraphics[width=0.14\textwidth,  ,valign=m, keepaspectratio,]{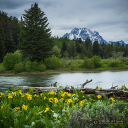} & 
\includegraphics[width=0.14\textwidth,  ,valign=m, keepaspectratio,]{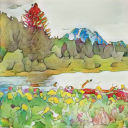}&
\includegraphics[width=0.14\textwidth,  ,valign=m, keepaspectratio,]{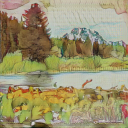}&
\includegraphics[width=0.14\textwidth,   ,valign=m, keepaspectratio,]{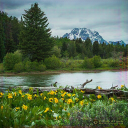} &
\includegraphics[width=0.14\textwidth,   ,valign=m, keepaspectratio,]{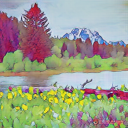}\\

{KH}  \includegraphics[width=0.14\textwidth,  ,valign=m, keepaspectratio,]{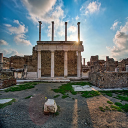} & 
\includegraphics[width=0.14\textwidth,  ,valign=m, keepaspectratio,]{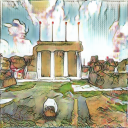}&
\includegraphics[width=0.14\textwidth,  ,valign=m, keepaspectratio,]{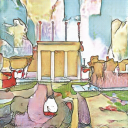}&
\includegraphics[width=0.14\textwidth,   ,valign=m, keepaspectratio,]{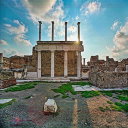} &
\includegraphics[width=0.14\textwidth,   ,valign=m, keepaspectratio,]{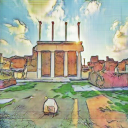}\\

{KP}  \includegraphics[width=0.14\textwidth,  ,valign=m, keepaspectratio,]{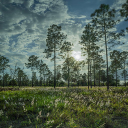}
& \includegraphics[width=0.14\textwidth,  ,valign=m, keepaspectratio,]{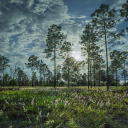}&
\includegraphics[width=0.14\textwidth,  ,valign=m, keepaspectratio,]{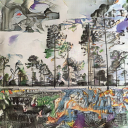}&
\includegraphics[width=0.14\textwidth,   ,valign=m, keepaspectratio,]{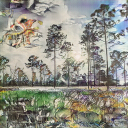}&
\includegraphics[width=0.14\textwidth,   ,valign=m, keepaspectratio,]{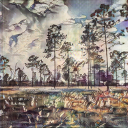} \\

{MB}  \includegraphics[width=0.14\textwidth,  ,valign=m, keepaspectratio,]{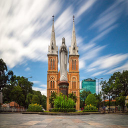} & 
\includegraphics[width=0.14\textwidth,  ,valign=m, keepaspectratio,]{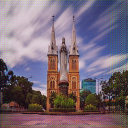}&
\includegraphics[width=0.14\textwidth,  ,valign=m, keepaspectratio,]{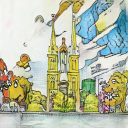}&
\includegraphics[width=0.14\textwidth,   ,valign=m, keepaspectratio,]{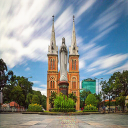} &
\includegraphics[width=0.14\textwidth,   ,valign=m, keepaspectratio,]{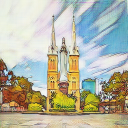}\\

{PP}  \includegraphics[width=0.14\textwidth,  ,valign=m, keepaspectratio,]{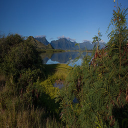} & 
\includegraphics[width=0.14\textwidth,  ,valign=m, keepaspectratio,]{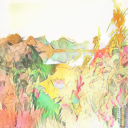}&
\includegraphics[width=0.14\textwidth,  ,valign=m, keepaspectratio,]{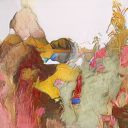}&
\includegraphics[width=0.14\textwidth,   ,valign=m, keepaspectratio,]{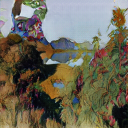} &
\includegraphics[width=0.14\textwidth,   ,valign=m, keepaspectratio,]{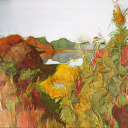} \\

{RC} \includegraphics[width=0.14\textwidth,  ,valign=m, keepaspectratio,]{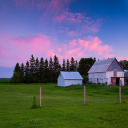} & 
\includegraphics[width=0.14\textwidth,  ,valign=m, keepaspectratio,]{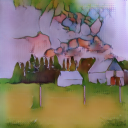}&
\includegraphics[width=0.14\textwidth,  ,valign=m, keepaspectratio,]{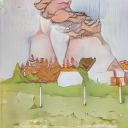}&
\includegraphics[width=0.14\textwidth,   ,valign=m, keepaspectratio,]{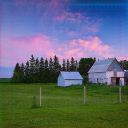}&
\includegraphics[width=0.14\textwidth,   ,valign=m, keepaspectratio,]{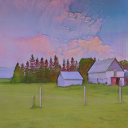}\\

{SC}  \includegraphics[width=0.14\textwidth,  ,valign=m, keepaspectratio,]{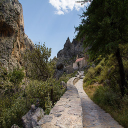} & 
\includegraphics[width=0.14\textwidth,  ,valign=m, keepaspectratio,]{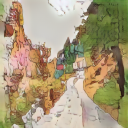}&
\includegraphics[width=0.14\textwidth,  ,valign=m, keepaspectratio,]{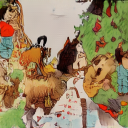}&
\includegraphics[width=0.14\textwidth,   ,valign=m, keepaspectratio,]{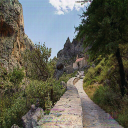} &
\includegraphics[width=0.14\textwidth,   ,valign=m, keepaspectratio,]{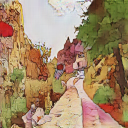}\\

{SD}  \includegraphics[width=0.14\textwidth,  ,valign=m, keepaspectratio,]{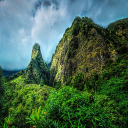} &
\includegraphics[width=0.14\textwidth,  ,valign=m, keepaspectratio,]{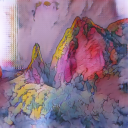}&
\includegraphics[width=0.14\textwidth,  ,valign=m, keepaspectratio,]{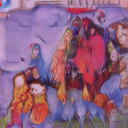}&
\includegraphics[width=0.14\textwidth,   ,valign=m, keepaspectratio,]{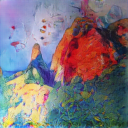}&
\includegraphics[width=0.14\textwidth,   ,valign=m, keepaspectratio,]{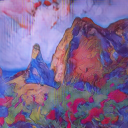}\\

{TR}  \includegraphics[width=0.14\textwidth,  ,valign=m, keepaspectratio,]{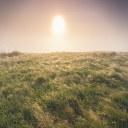} & 
\includegraphics[width=0.14\textwidth,  ,valign=m, keepaspectratio,]{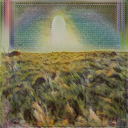}&
\includegraphics[width=0.14\textwidth,  ,valign=m, keepaspectratio,]{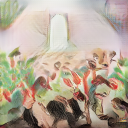}&
\includegraphics[width=0.14\textwidth,   ,valign=m, keepaspectratio,]{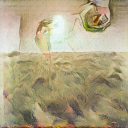} &
\includegraphics[width=0.14\textwidth,   ,valign=m, keepaspectratio,]{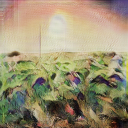}\\

\end{tabular}
\vspace{-3mm}
\caption{
Sample results from CartoonGAN, CycleGAN, DualGAN and GANILLA. Input images on each row are transferred with a different style for variety. Most of the time, CycleGAN discards the content (see RC, TR), whereas DualGAN fails to transfer the style at all (see DM, KH and SC).   GANILLA transfers the style and preserves the content better than the other models.
\label{fig:all-results}}
\end{figure*}


\subsection{Qualitative Analysis and User Study}
\label{sec:qualitative}

We present style transfer results of GANILLA in Figure~\ref{fig:our-results}. Using same input test image, GANILLA generates images with the style of unique artist. Although for some illustrators, generated images contain unobtrusive defects, most of them captures target style successfully.

Figure~\ref{fig:all-results} presents sample outputs from four methods for different styles. Although CycleGAN captures the style of the illustrators well, it is hard to tell what the content is. Also it hallucinates things such as faces and objects from source illustrations, on the generated images. On the other hand, CartoonGAN and DualGAN preserve content but in many examples, they lack in transferring the style. GANILLA successfully generates stylized images while preserving content.

\begin{figure*}
\captionsetup[subfigure]{labelformat=empty}
\centering
\setlength\tabcolsep{1.5pt} 
\begin{tabular}{cc}
\includegraphics[width=0.20\textwidth,  ,valign=t, keepaspectratio,]{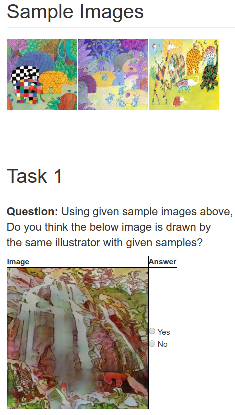} & 
\includegraphics[width=0.65\textwidth,  ,valign=t, keepaspectratio,]{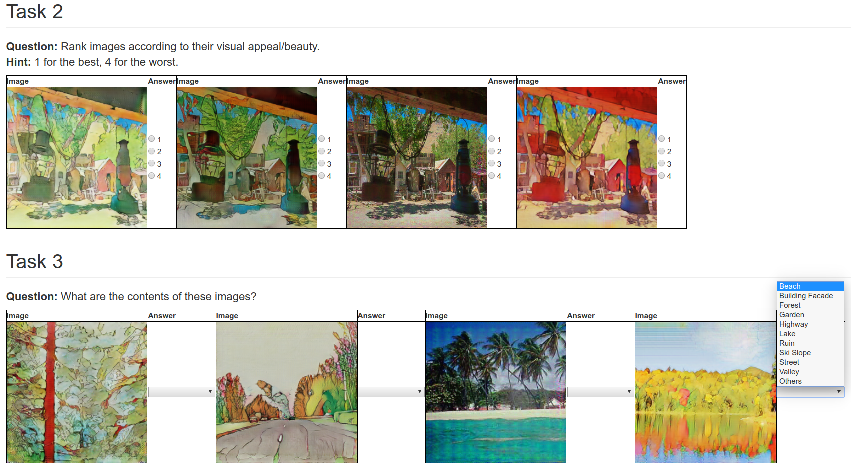}

\end{tabular}
\vspace{-3mm}
\caption{Sample screenshots from our user study web page.}
\label{fig:ss-user-study}
\end{figure*}

Visual inspection based qualitative analysis is subjective. To reduce the effect of subjectiveness, we conducted a user study~\footnote{Readers could reach our study at http://18.217.61.29}. In this study, the user is presented with a web page (see a sample in Figure~\ref{fig:ss-user-study}) containing sample illustrations from an artist and three different questions to answer. The artist of the sample illustrations is not disclosed to the user. For the first task, the user is asked a Yes/No question: “Was this image drawn by the artist of the sample illustrations?” For the second task, we ask the user to rank the given four images according to their visual appeal from 1 to 4, where 1 is for the most appealing one. These four images belong to four methods (CycleGAN, CartoonGAN, DualGAN and GANILLA), and they are placed randomly. 
For the final task, we ask the user to identify the content categories of the given images. This is a multiple choice question. The user tries to select the correct content from ten different outdoor categories such as beach, forest and valley.

\begin{table}[h]
\begin{center}
 \caption{User Study results.}
 \label{tab:user-study}
 \resizebox{0.7\columnwidth}{!}{
 \begin{tabular}{lcccc}
   \toprule 
   & CartoonGAN & CycleGAN & DualGAN & GANILLA         \\
   \cmidrule(lr){2-5}
   Style                 & 40.9         & 44.4         & 54.5                 &  \bf 60.0        \\
   Content                 & 67.0         & 69.4        & 68.1          &  \bf 70.0        \\
   Content \& Style      & 54.0 & 56.9 & 61.3                 &  \bf 65.0         \\ 
 \cmidrule(lr){2-5}
   Avg. Ranking         & 2.68         & 2.89        & \bf  2.18         &  2.21          \\
   \bottomrule 
 \end{tabular}}
\end{center}
\end{table}

We collected a total of 66 survey inputs from 48 different users. 
In Table~\ref{tab:user-study}, we present user study results in terms of average accuracies for Style and Content detection tasks, and average ranking for visually appealing task.

In terms of style and content identifiability, GANILLA appeared to be the best method. Content evaluation (second row in Table~\ref{tab:user-study}) scores are similar across models, which show that the users are able to find the content successfully.

In the visually appealingness task, DualGAN exceeds GANILLA with a slight margin (last row in Table~\ref{tab:user-study}). The main reason for that is DualGAN fails to transfer style for some illustrators, and users generally pick these images as visually appealing since they look more natural.

\begin{table*}[h]
\begin{center}
\caption{Comparison of Style-CNN and Content-CNN results.}
\label{tab:cnn-results}
\resizebox{1.\columnwidth}{!}{
\begin{tabular}{lcccc:cccc:cccc}
 \toprule 
 \multirow{2}{*}{Style} & \multicolumn{4}{c}{Content CNN (\%)} & \multicolumn{4}{c}{Style CNN (\%)} & \multicolumn{4}{c}{Final Scores (\%)} \\
 \cmidrule(lr){2-5}  \cmidrule(lr){6-9}   \cmidrule(lr){10-13}
& \shortstack{Cartoon\\GAN} & \shortstack{Cycle\\GAN} & \shortstack{Dual\\GAN} & \shortstack{GAN \\ ILLA} & \shortstack{Cartoon\\GAN} &  \shortstack{Cycle\\GAN} &  \shortstack{Dual\\GAN} & \shortstack{GAN \\ ILLA}  & \shortstack{Cartoon\\GAN} &  \shortstack{Cycle\\GAN} &  \shortstack{Dual\\GAN} & \shortstack{GAN \\ ILLA}        \\
 \midrule 
AS         & 16.4         & 9.6         & 33.2        &\textbf{41.2}        & 77.8 & \textbf{99.2} & 94.8 & 95.3        & 47.1         & 54.4 & 64.0 & \textbf{68.3}        \\
DM         & 1.0         & 0.4         &  \textbf{87.0}& 11.0         & 64.4 & \textbf{88.8} & 0.5 & 81.4                 & 32.7         & 44.6 & 43.8 & \textbf{46.2}        \\
KH         & 0.4         & 1.2         & \textbf{87.4}        &   9.5        & 87.1 & \textbf{99.0} & 0.2 & 81.6                 & 43.8        & \textbf{50.1} & 43.8 & 45.6        \\
KP         & \textbf{89.6}        & 15.4  & 31.0        &   26.6         & 0.7        & \textbf{99.2} & 94.1 & 89.6                & 45.2 & 57.3 & \textbf{62.6} & 58.1         \\
MB         & 85.6         & 4.4         & \textbf{87.4}        & 15.0                 & 0.1 & \textbf{94.0} & 1.4 & 67.0                 & 42.9         & \textbf{49.2} & 44.4 & 41.0        \\
PP         & 8.2   & 10.2  & \textbf{36.0}        &   26.8         & 85.0& \textbf{96.4} & 86.1 & 91.7                 & 46.6         & 53.3 & \textbf{61.1} & 59.3        \\
RC         & 1.0   & 0.4          & \textbf{87.4} &   25.4         & 98.9         & \textbf{99.8} & 0.1  & 85.6                & 50.0         & 50.1 & 43.8 & \textbf{55.5}         \\
SC         & 3.8         & 2.6         & \textbf{85.2}        &   18.6         & 92.8 & \textbf{99.8} & 1.3 & 59.4                & 48.3         & \textbf{51.2} & 43.3 & 39.0        \\
SD         & 3.8        & 1.0   & \textbf{5.8}         &   4.8          & 68.0         & 96.6 & 94.0 & \textbf{98.4}                & 35.9         & 48.8 & 49.9 & \textbf{51.6} \\
 TR         & 22.0        & 6.6          & \textbf{34.4}        &   20.6         & 30.9         & \textbf{99.3} & 92.0 & 94.7                & 26.5         & 53.0 & \textbf{63.2} & 57.7 \\
\hline
Avg. & 23.2 & 5.2  & \textbf{57.5} & 20.0     & 60.6 & \textbf{97.2} & 46.5 & 84.5    &  41.9 &         51.2 &         52.0 &   \textbf{52.2}                \\
 \bottomrule 
\end{tabular}}
\end{center}
\end{table*}

\subsection{Quantitative Analysis}
\label{sec:quantitative}

\begin{figure}
\centering
\begin{tabular}{rcccc}
 & \shortstack{High Style \\ Error} \hspace*{-12pt}  & \shortstack{Low Style \\ Error}  \hspace*{-12pt} & \shortstack{Correct w \\ Low Conf}  \hspace*{-12pt} & \shortstack{Correct w \\ High Conf}  \hspace*{-12pt} \\
\shortstack{Input} \hspace*{-12pt} & \includegraphics[width=0.09\textwidth,  ,valign=m, keepaspectratio,]{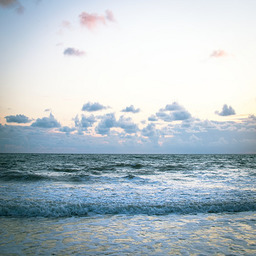}  \hspace*{-12pt} & 
\includegraphics[width=0.09\textwidth,  ,valign=m, keepaspectratio,]{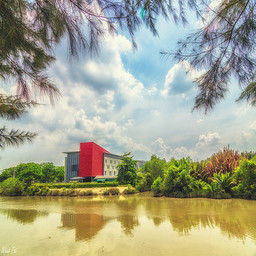} \hspace*{-12pt} &
\includegraphics[width=0.09\textwidth,  ,valign=m, keepaspectratio,]{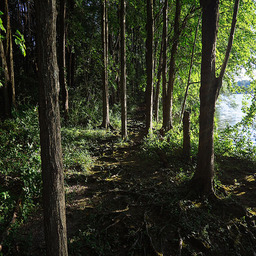} \hspace*{-12pt} &
\includegraphics[width=0.09\textwidth,   ,valign=m, keepaspectratio,]{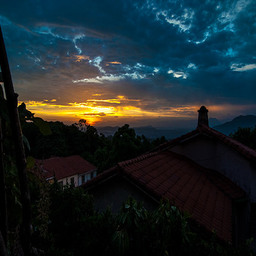} \hspace*{-12pt} \\

\shortstack{Target \\ style:  PP}  \hspace*{-12pt} & \includegraphics[width=0.09\textwidth,  ,valign=m, keepaspectratio,]{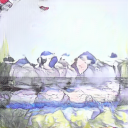}  \hspace*{-12pt}& 
\includegraphics[width=0.09\textwidth,  ,valign=m, keepaspectratio,]{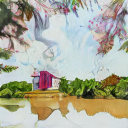} \hspace*{-12pt}&
\includegraphics[width=0.09\textwidth,  ,valign=m, keepaspectratio,]{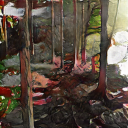} \hspace*{-12pt}&
\includegraphics[width=0.09\textwidth,   ,valign=m, keepaspectratio,]{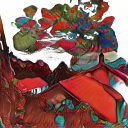} \hspace*{-12pt} \\

\midrule
\shortstack{Input}  \hspace*{-12pt} & \includegraphics[width=0.09\textwidth,  ,valign=m, keepaspectratio,]{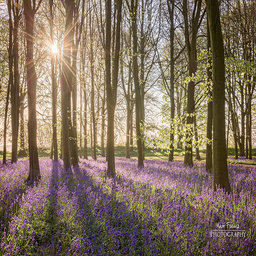}  \hspace*{-12pt}& 
\includegraphics[width=0.09\textwidth,  ,valign=m, keepaspectratio,]{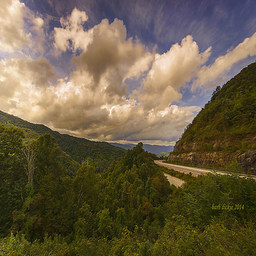} \hspace*{-12pt}&
\includegraphics[width=0.09\textwidth,  ,valign=m, keepaspectratio,]{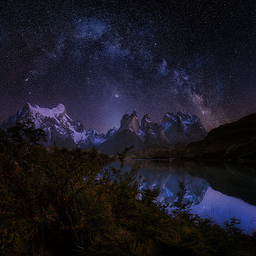} \hspace*{-12pt}&
\includegraphics[width=0.09\textwidth,   ,valign=m, keepaspectratio,]{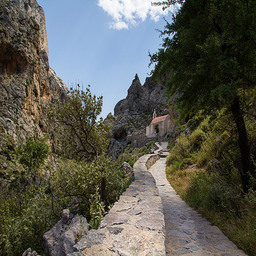} \hspace*{-12pt}\\

\shortstack{Target \\ style:  AS}  \hspace*{-12pt} & \includegraphics[width=0.09\textwidth,  ,valign=m, keepaspectratio,]{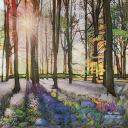}  \hspace*{-12pt}& 
\includegraphics[width=0.09\textwidth,  ,valign=m, keepaspectratio,]{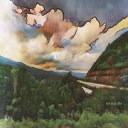} \hspace*{-12pt}&
\includegraphics[width=0.09\textwidth,  ,valign=m, keepaspectratio,]{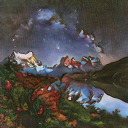} \hspace*{-12pt}&
\includegraphics[width=0.09\textwidth,   ,valign=m, keepaspectratio,]{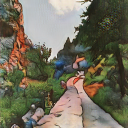} \hspace*{-12pt}\\
\end{tabular}
\caption{Sample evaluations from Style-CNN. The first two rows show translation examples by CycleGAN and the last two rows are examples by GANILLA. From left to right, columns represent increasing levels of style accuracy. ``High Style Error” means that the generated image for the target style is misclassified with high confidence score. Similarly, ``Low Style Error”: misclassified with low confidence score. ``Correct w Low Conf”: classified correctly but with low confidence score. ``Correct w High Conf”: classified correctly with a high confidence score. These examples qualitatively validates our Style-CNN’s decisions.}
\label{fig:quant_fig_analyse_stil}
\end{figure}

To quantitatively evaluate the quality of style transfer, we propose to use a style classifier. In order to train a style specific classifier, we must detach the training images from their visual content while keeping the style in them. For this purpose, we randomly cropped small patches (i.e $100\times100$ pixel patches) from illustration images and used these patches to train our style classifier, the \textit{Style-CNN}. Our training set had 11 classes: 10 for the illustration artists and one for natural images. Our intuition in adding the 11th class is that, if a generated image lacks style, then it is more likely to be classified as a natural image. We used only generated images to test the classifier. Figure \ref{fig:quant_fig_analyse_stil} presents some translation examples from CycleGAN and GANILLA to show that Style-CNN decisions are justifiable.


\newcommand{\rulesep}{\unskip\ \hrule\ }

\begin{figure}
\centering
\begin{tabular}{rcccc}
& \shortstack{High Cont. \\ Error} \hspace*{-12pt}  & \shortstack{Low Cont. \\ Error}  \hspace*{-12pt} & \shortstack{Correct w \\ Low Conf}  \hspace*{-12pt} & \shortstack{Correct w \\ High Conf}  \hspace*{-12pt} \\
\shortstack{Input} \hspace*{-12pt} & \includegraphics[width=0.09\textwidth,  ,valign=m, keepaspectratio,]{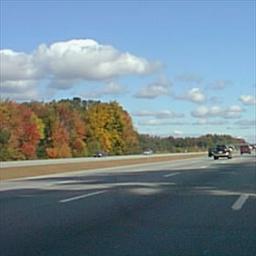} \hspace*{-12pt} & 
\includegraphics[width=0.09\textwidth,  ,valign=m, keepaspectratio,]{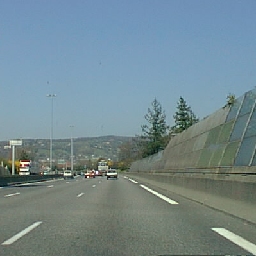} \hspace*{-12pt} &
\includegraphics[width=0.09\textwidth,  ,valign=m, keepaspectratio,]{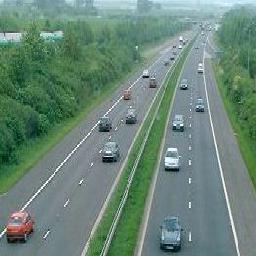} \hspace*{-12pt} &
\includegraphics[width=0.09\textwidth,   ,valign=m, keepaspectratio,]{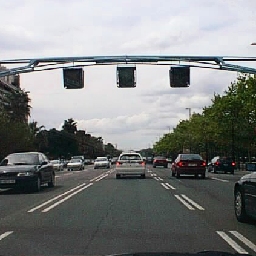}  \hspace*{-12pt} \\

\shortstack{Target \\ Style: KP}  \hspace*{-12pt} & \includegraphics[width=0.09\textwidth,  ,valign=m, keepaspectratio,]{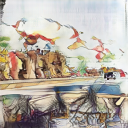}  \hspace*{-12pt}& 
\includegraphics[width=0.09\textwidth,  ,valign=m, keepaspectratio,]{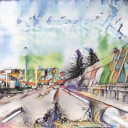} \hspace*{-12pt}&
\includegraphics[width=0.09\textwidth,  ,valign=m, keepaspectratio,]{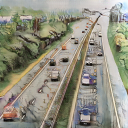} \hspace*{-12pt}&
\includegraphics[width=0.09\textwidth,   ,valign=m, keepaspectratio,]{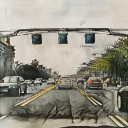} \hspace*{-12pt} \\

\midrule

\shortstack{Input}  \hspace*{-12pt} & \includegraphics[width=0.09\textwidth,  ,valign=m, keepaspectratio,]{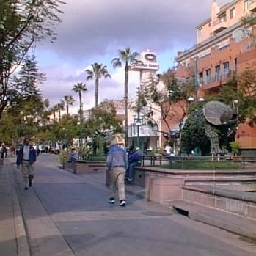}  \hspace*{-12pt} & 
\includegraphics[width=0.09\textwidth,  ,valign=m, keepaspectratio,]{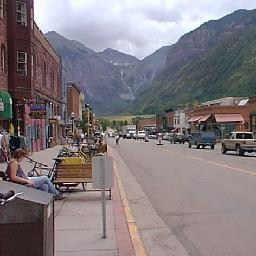} \hspace*{-12pt}&
\includegraphics[width=0.09\textwidth,  ,valign=m, keepaspectratio,]{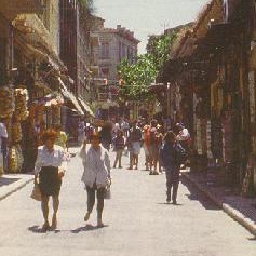} \hspace*{-12pt}&
\includegraphics[width=0.09\textwidth,   ,valign=m, keepaspectratio,]{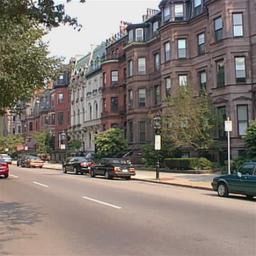} \hspace*{-12pt}\\

\shortstack{Target \\ Style: PP}  \hspace*{-12pt} & \includegraphics[width=0.09\textwidth,  ,valign=m, keepaspectratio,]{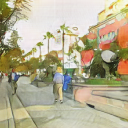}  \hspace*{-12pt}& 
\includegraphics[width=0.09\textwidth,  ,valign=m, keepaspectratio,]{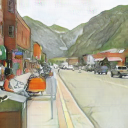} \hspace*{-12pt}&
\includegraphics[width=0.09\textwidth,  ,valign=m, keepaspectratio,]{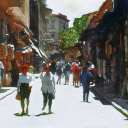} \hspace*{-12pt}&
\includegraphics[width=0.09\textwidth,   ,valign=m, keepaspectratio,]{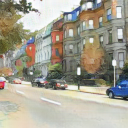} \hspace*{-12pt}\\
\end{tabular}
\caption{Sample evaluations from Content-CNN. The first two rows show translation examples by CycleGAN and the last two rows are examples by GANILLA. From left to right, columns represent increasing levels of content accuracy. ``High Content Error” means that the generated image is misclassified with high confidence score. Similarly, ``Low Content Error”: misclassified with low confidence score. ``Correct w Low Conf”: classified correctly but with low confidence score. ``Correct w High Conf”: classified correctly with a high confidence score. }
\label{fig:quant_fig_analyse_content}
\end{figure}

We argue that capturing the style is not sufficient if the content of the source is not preserved. For evaluating the content preservation, we propose a content classifier~\textit{Content-CNN}. Here we define content as belonging to a specific scene category (e.g.  forest, street, etc.). We select ten outdoor classes, which are close to the content of natural image dataset, from SUN dataset~\cite{SUN}. 4150 training and 500 testing images are used. 
As the negative class, we use the full illustrators dataset excluding the images corresponding to ten illustrators used in training. Our intuition is that, if we are able to preserve the content, then a natural image that is stylized in an illustrator’s style should still have the same content. For example, if we generate a mountain image in Korky Paul style, we should still be able to classify it as a mountain. If the generated image has lost its connection with the content, then it could be classified as an illustration, i.e. in the negative class. If the content is not preserved, it will be just an illustration with no specific definition of the scene. Figure~\ref{fig:quant_fig_analyse_content} presents some translation examples from CycleGAN and GANILLA to show that Content-CNN decisions are justifiable.

\begin{table}[h]
\begin{center}
 \caption{Style\&Content CNN results of Ablation Models.}
 \label{tab:cnn-results-abl}
 \resizebox{0.7\columnwidth}{!}{
 \begin{tabular}{lcc:cc:cc}
   \toprule 
   \multirow{2}{*}{Style} & \multicolumn{2}{c}{Content CNN (\%)} & \multicolumn{2}{c}{Style CNN (\%)}  & \multicolumn{2}{c}{Final Scores (\%)} \\
   \cmidrule(lr){2-3}        \cmidrule(lr){4-5}   \cmidrule(lr){6-7}
   & Model 1 & Model 2 & Model 1 & Model 2  & Model 1 & Model 2\\
   \midrule 
   AS         &        61.6        &         1.6  &        73.7         &         98.0         &        67.7         &         49.8        \\
   KP         &         22.6        &         1.2  &         90.5        &         99.2          &         56.6        &         50.2        \\
   PP         &         18.0        &         1.2  &         90.1        &         95.7        &         54.1        &         48.5        \\
   RC         &         3.6         &         0.2  &         88.4        &         98.6         &         46.0        &         49.4        \\
   SD         &         9.0         &         0.0  &         93.6        &          88.1          &         51.3        &         44.1        \\
\hline
Avg.           &         23.4         &   0.7        &         88.1         &         96.4         &         55.7        &    48.5        \\    \bottomrule 
 \end{tabular}}
\end{center}
\end{table}

Style-CNN and Content-CNN results are given in Table~\ref{tab:cnn-results}. We present classification accuracies for each illustrator separately and also provide the average of all results. To calculate a final score for all methods, we average content and style scores, given in the last row. GANILLA achieves best overall score compared to others. CycleGAN outperforms others in style, while DualGAN got the best content score. Since CycleGAN ``hallucinates" random crops from input illustration images such as animal or person faces, its style accuracy is higher than other methods. On the other hand, CartoonGAN and DualGAN failed to learn specific styles (KP, MB and DM, KH, MB, RC, SC respectively), and their style score is lower than others. When the style is not transferred, the content is likely to be preserved, thus their content score is higher. Except SD, GANILLA gives consistently high scores for both content and style.

\begin{figure*}[t]
\captionsetup[subfigure]{labelformat=empty}
\centering
\begin{subfigure}[b]{0.13\textwidth}
    \caption{Input}
\includegraphics[width=1.0\textwidth]{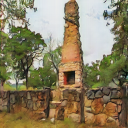}
\vspace{0.05 cm}
\includegraphics[width=1.0\textwidth]{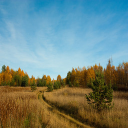}
\vspace{0.05 cm}
\includegraphics[width=1.0\textwidth]{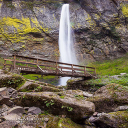}
\end{subfigure}
\begin{subfigure}[b]{0.13\textwidth}
    \caption{\shortstack{Model 1 \\ AS}}
\includegraphics[width=1.0\textwidth]{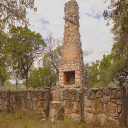}
\vspace{0.05 cm}
\includegraphics[width=1.0\textwidth]{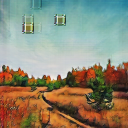}
\vspace{0.05 cm}
\includegraphics[width=1.0\textwidth]{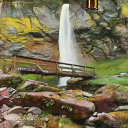}
\end{subfigure}
\begin{subfigure}[b]{0.13\textwidth}
    \caption{\shortstack{Model 2 \\ AS}}
\includegraphics[width=1.0\textwidth]{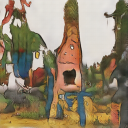}
\vspace{0.05 cm}
\includegraphics[width=1.0\textwidth]{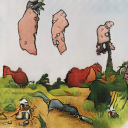}
\vspace{0.05 cm}
\includegraphics[width=1.0\textwidth]{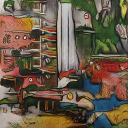}
\end{subfigure}
\begin{subfigure}[b]{0.13\textwidth}
    \caption{\shortstack{GANILLA \\ AS}}
\includegraphics[width=1.0\textwidth]{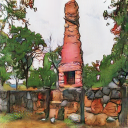}
\vspace{0.05 cm}
\includegraphics[width=1.0\textwidth]{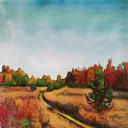}
\vspace{0.05 cm}
\includegraphics[width=1.0\textwidth]{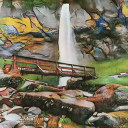}
\end{subfigure}
\begin{subfigure}[b]{0.13\textwidth}
    \caption{\shortstack{Model 1 \\ KP}}
\includegraphics[width=1.0\textwidth]{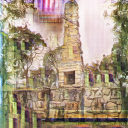}
\vspace{0.05 cm}
\includegraphics[width=1.0\textwidth]{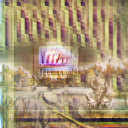}
\vspace{0.05 cm}
\includegraphics[width=1.0\textwidth]{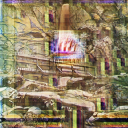}
\end{subfigure}
\begin{subfigure}[b]{0.13\textwidth}
    \caption{\shortstack{Model 2 \\ KP}}
\includegraphics[width=1.0\textwidth]{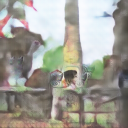}
\vspace{0.05 cm}
\includegraphics[width=1.0\textwidth]{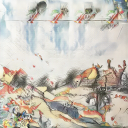}
\vspace{0.05 cm}
\includegraphics[width=1.0\textwidth]{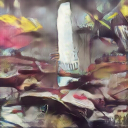}
\end{subfigure}
\begin{subfigure}[b]{0.13\textwidth}
    \caption{\shortstack{GANILLA \\ KP}}
\includegraphics[width=1.0\textwidth]{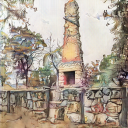}
\vspace{0.05 cm}
\includegraphics[width=1.0\textwidth]{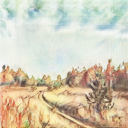}
\vspace{0.05 cm}
\includegraphics[width=1.0\textwidth]{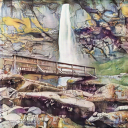}
\end{subfigure}
\caption{For styles AS and KP, results of ablation models (Model1 and Model2) are compared with GANILLA’s results. }
\label{fig:abl_exps}
\end{figure*}

\subsection{Ablation Experiments}

In order to evaluate the effects of different parts of our model in detail, we conducted two ablation experiments. It is possible to partition our model into two parts, the first one is downsampling part and the second one is upsampling part. In our first ablation experiment, we replaced our downsampling CNN with the original ResNet-18 to see the effects of modifications.

Our second ablation model is composed by using our downsampling CNN with deconv layers for upsampling part similar to CycleGAN and CartoonGAN. In this model, output of our downsampling CNN (only last layer) is fed to a series of deconvolution layers. This model aims to measure the effects of using multiple feature layers for the upsampling part.

In our ablation experiments, we trained models with the same ten illustrator styles, but present results for randomly selected five of them since other results are inline. Visual results of all the ablation experiments are given in Figure~\ref{fig:abl_exps}. In Table~\ref{tab:cnn-results-abl} we present style and content classifier results for ablation experiments.

Ablation Model~1 gives similar content score with GANILLA, but its style score is lower. This shows that our modifications to original ResNet-18 architecture enabled GANILLA to successfully stylize input images. On the other hand, Ablation Model~2 achieves better style score than GANILLA but its content score is too low. This demonstrates that using low level features in the up-sampling part helps to preserve content.

\begin{figure*}[h]
\captionsetup[subfigure]{labelformat=empty}
\centering
\begin{subfigure}[b]{0.16\textwidth}
\centering
  \includegraphics[width=1.0\textwidth]{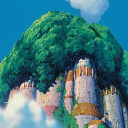}
  \vspace{0.05 cm}
  \includegraphics[width=1.0\textwidth]{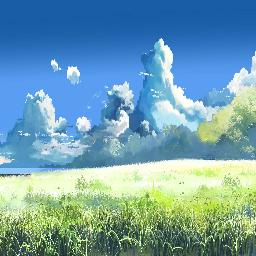}
\end{subfigure}
\begin{subfigure}[b]{0.16\textwidth}
\centering
  \includegraphics[width=1.0\textwidth]{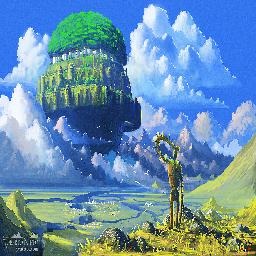}
  \vspace{0.05 cm}
  \includegraphics[width=1.0\textwidth]{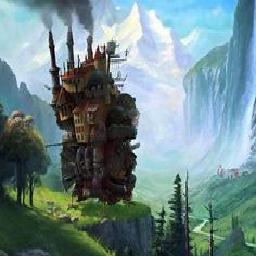}
\end{subfigure}
\begin{subfigure}[b]{0.16\textwidth}
\centering
  \includegraphics[width=1.0\textwidth]{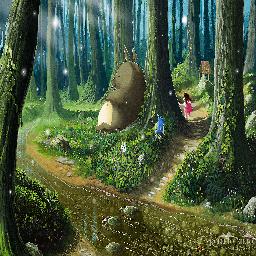}
  \vspace{0.05 cm}
  \includegraphics[width=1.0\textwidth]{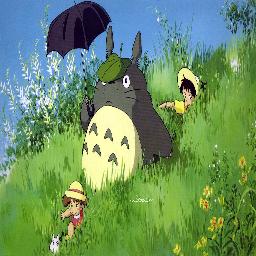}
\end{subfigure}
\begin{subfigure}[b]{0.16\textwidth}
\centering
  \includegraphics[width=1.0\textwidth]{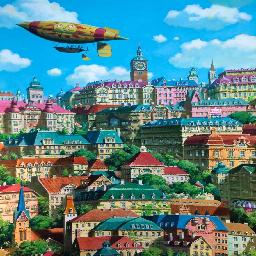}
  \vspace{0.05 cm}
 \includegraphics[width=1.0\textwidth]{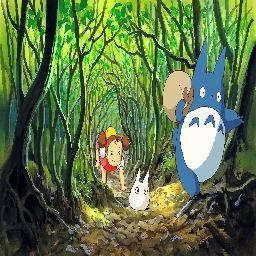}
\end{subfigure}
\begin{subfigure}[b]{0.16\textwidth}
\centering
  \includegraphics[width=1.0\textwidth]{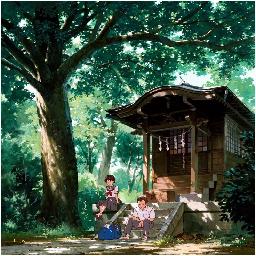}
  \vspace{0.05 cm}
  \includegraphics[width=1.0\textwidth]{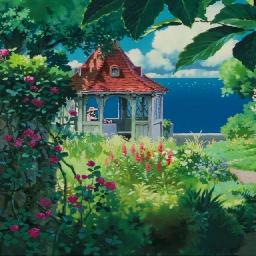}
\end{subfigure}
\begin{subfigure}[b]{0.16\textwidth}
\centering
  \includegraphics[width=1.0\textwidth]{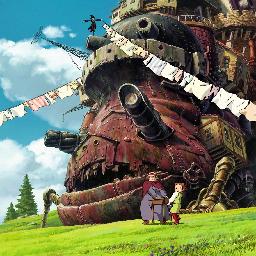}
  \vspace{0.05 cm}
  \includegraphics[width=1.0\textwidth]{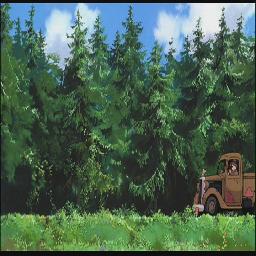}
\end{subfigure}
\caption{Example images from collected Hayao dataset.}
\label{fig:sample-hayao}
\end{figure*}

\subsection{Cartoonization Using Hayao Style}

In~\cite{cartoongan}, dataset is constructed by sampling images from the videos of stories from Miyazaki Hayao. Since, the dataset was not publicly available we have tried repeating the same procedure. However, we were not able to replicate the results of CartoonGAN on our collection due to the low quality of the samples. In order to be able to perform experiments on the same illustrator with CartoonGAN, instead, we have collected images of Miyazaki Hayao using Google Image Search to be used as our target training set. Please note that, this is a more challenging dataset compared to the Hayao dataset used in \cite{cartoongan} which was composed of a unique style from a single story (Spirited Away). Our images correspond to the samples from the entire collection of Hayao and therefore mixture of several styles from a variety of stories.

In Figure~\ref{fig:sample-hayao}, we show samples from our Hayao collection. Note that, compared to the illustrator dataset presented in our study, abstraction in Hayao dataset is very limited. Illustrations of Hayao mostly correspond to natural scenery, and therefore are more similar to the input images used in the tests.

\begin{figure*}
\captionsetup[subfigure]{labelformat=empty}
\centering
\begin{subfigure}[b]{0.16\textwidth}
 \caption{Input}
\includegraphics[width=1.0\textwidth]{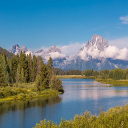}
\vspace{0.05 cm}
\includegraphics[width=1.0\textwidth]{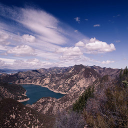}
\vspace{0.05 cm}
\includegraphics[width=1.0\textwidth]{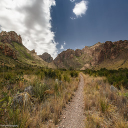}
\end{subfigure}
\begin{subfigure}[b]{0.16\textwidth}
 \caption{Output}
\includegraphics[width=1.0\textwidth]{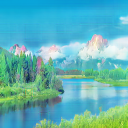}
\vspace{0.05 cm}
\includegraphics[width=1.0\textwidth]{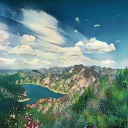}
\vspace{0.05 cm}
\includegraphics[width=1.0\textwidth]{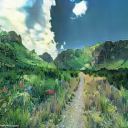}
\end{subfigure}
\begin{subfigure}[b]{0.16\textwidth}
 \caption{Input}
\includegraphics[width=1.0\textwidth]{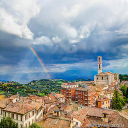}
\vspace{0.05 cm}
\includegraphics[width=1.0\textwidth]{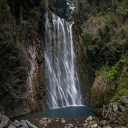}
\vspace{0.05 cm}
\includegraphics[width=1.0\textwidth]{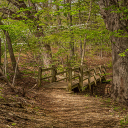}
\end{subfigure}
\begin{subfigure}[b]{0.16\textwidth}
 \caption{Output}
\includegraphics[width=1.0\textwidth]{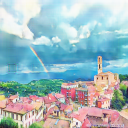}
\vspace{0.05 cm}
\includegraphics[width=1.0\textwidth]{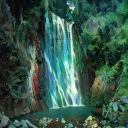}
\vspace{0.05 cm}
\includegraphics[width=1.0\textwidth]{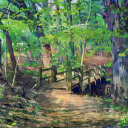}
\end{subfigure}
\begin{subfigure}[b]{0.16\textwidth}
 \caption{Input}
\includegraphics[width=1.0\textwidth]{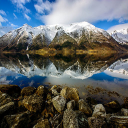}
\vspace{0.05 cm}
\includegraphics[width=1.0\textwidth]{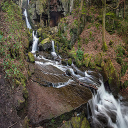}
\vspace{0.05 cm}
\includegraphics[width=1.0\textwidth]{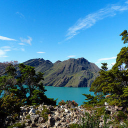}
\end{subfigure}
\begin{subfigure}[b]{0.16\textwidth}
 \caption{Output}
\includegraphics[width=1.0\textwidth]{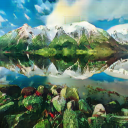}
\vspace{0.05 cm}
\includegraphics[width=1.0\textwidth]{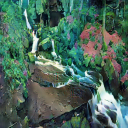}
\vspace{0.05 cm}
\includegraphics[width=1.0\textwidth]{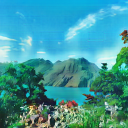}
\end{subfigure}
\caption{Sample results of GANILLA using Hayao as style.}
\label{fig:hayao-results}
\end{figure*}

We only present visual results for the cartoonization and the results of our method, GANILLA, shown in Figure~\ref{fig:hayao-results}. The results show the effectiveness of GANILLA in the cartoonization domain. Especially, since the color green is dominant in source images, transferred images also contain greenish background.

\subsection{Comparisons with Neural Style Transfer}

\begin{figure*}[h]
\captionsetup[subfigure]{labelformat=empty}
\centering
\setlength\tabcolsep{1.5pt} 
\begin{tabular}{rccc}
{Content Image}  & {Style Image}  & Gatys~\etal & GANILLA \\
{AS}  \includegraphics[width=0.15\textwidth, ,valign=m, keepaspectratio,]{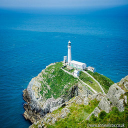} &
\includegraphics[width=0.15\textwidth, ,valign=m, keepaspectratio,]{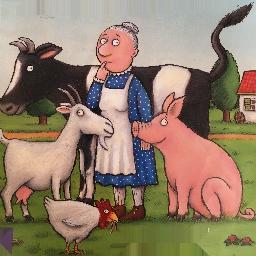} &
\includegraphics[width=0.15\textwidth, ,valign=m, keepaspectratio,]{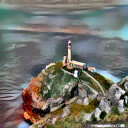}&
\includegraphics[width=0.15\textwidth, ,valign=m, keepaspectratio,]{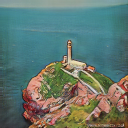}\\

{KP}  \includegraphics[width=0.15\textwidth, ,valign=m, keepaspectratio,]{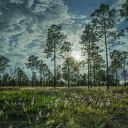} &
\includegraphics[width=0.15\textwidth, ,valign=m, keepaspectratio,]{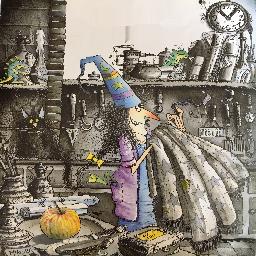} &
\includegraphics[width=0.15\textwidth, ,valign=m, keepaspectratio,]{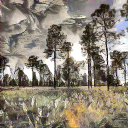}&

\includegraphics[width=0.15\textwidth, ,valign=m, keepaspectratio,]{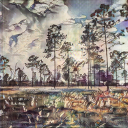}\\

{PP}  \includegraphics[width=0.15\textwidth, ,valign=m, keepaspectratio,]{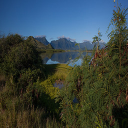} &
\includegraphics[width=0.15\textwidth, ,valign=m, keepaspectratio,]{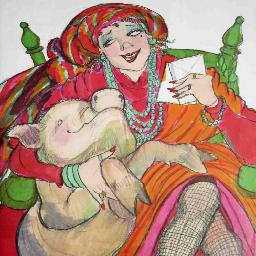} &
\includegraphics[width=0.15\textwidth, ,valign=m, keepaspectratio,]{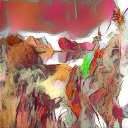}&

\includegraphics[width=0.15\textwidth, ,valign=m, keepaspectratio,]{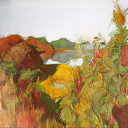}\\

{RC}  \includegraphics[width=0.15\textwidth, ,valign=m, keepaspectratio,]{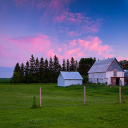} &
\includegraphics[width=0.15\textwidth, ,valign=m, keepaspectratio,]{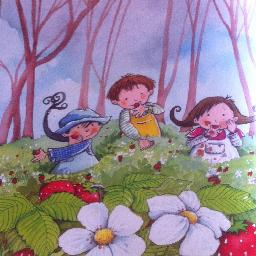} &
\includegraphics[width=0.15\textwidth, ,valign=m, keepaspectratio,]{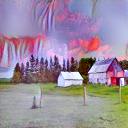}&
\includegraphics[width=0.15\textwidth, ,valign=m, keepaspectratio,]{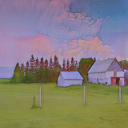}\\

\end{tabular}
\caption{
Comparison with Neural Style Transfer (NST) method~\cite{gatsy2} on photo stylization. Each row presents results of one style using~\cite{gatsy2} and GANILLA. First column shows input content images, second column presents style images used for~\cite{gatsy2}. Third column give results of~\cite{gatsy2} and finally, last column presents results of GANILLA. 
\label{fig:nst-compare-1}}
\end{figure*}

We compare our method with Gatys~\etal~\cite{gatsy2} NST method in Figure~\ref{fig:nst-compare-1}. For Gatys~\etal NST model, we use one content and one style image. As it could be seen from the figure, NST method is not successful in terms of stylization on illustration dataset. Since NST uses only one source and style image, its result is highly depends on the selected input images. Although NST transfers color information correctly, it falls short of dealing with the style.

\subsection{Limitations and Discussion}

\begin{figure}
\captionsetup[subfigure]{labelformat=empty}
\centering
\begin{subfigure}[b]{0.15\textwidth}
    \caption{Input}
\includegraphics[width=1.0\textwidth]{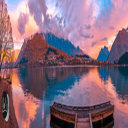}
\vspace{0.02 cm}
\includegraphics[width=1.0\textwidth]{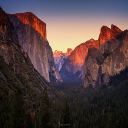}
\end{subfigure}
\begin{subfigure}[b]{0.15\textwidth}
    \caption{Output}
\includegraphics[width=1.0\textwidth]{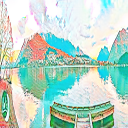}
\vspace{0.02 cm}
\includegraphics[width=1.0\textwidth]{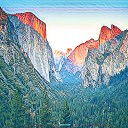}
\end{subfigure}
\begin{subfigure}[b]{0.15\textwidth}
    \caption{Input}
\includegraphics[width=1.0\textwidth]{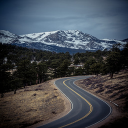}
\vspace{0.02 cm}
\includegraphics[width=1.0\textwidth]{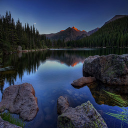}
\end{subfigure}
\begin{subfigure}[b]{0.15\textwidth}
    \caption{Output}
\includegraphics[width=1.0\textwidth]{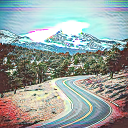}
\vspace{0.02 cm}
\includegraphics[width=1.0\textwidth]{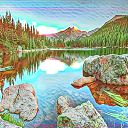}
\end{subfigure}
\vspace{-10pt}
\caption{Example failure cases of GANILLA.}
\label{fig:fail_cases}
\end{figure}

Our model fails for some illustrators in terms of stylization. In Figure~\ref{fig:fail_cases}, we provide an example case where style illustrator is Dr. Seuss. Main reason for this failure is that Dr. Seuss’ illustrations are mostly charcoal drawings or contain simple colorings.

Here, we would like to note that the illustration dataset presented in this study has several challenges compared to the datasets used in other style transfer studies. 
For example, for the dataset used in CartoonGAN, each style is obtained from the images of a single video of a story \footnote{We were not able to use the datasets in CartoonGAN since it was not publicly available}. On the other hand, the illustration dataset includes variety of books for each illustrator, and the style may change drastically from one book to another. Moreover, illustrations mostly include characters (animals or people) which are very different from the natural scenery images. As observed in Dr. Seuss illustrations the white/plain backgrounds and sharp boundaries also cause difficulties.

\section{Conclusions}

In this paper, we presented the most extensive children’s books illustration dataset and a new generator network for image-to-illustration translation. Since children’s book illustrations contain highly abstract objects and shapes, current state-of-the-art generator networks fail to transfer the content and style at the same time. To overcome this issue, our model uses low level features in downsampling state as well as in the upsampling part.

One major problem in the image-to-image translation domain is that there are no well defined or agreed-upon metrics to evaluate a generator model. To address this problem, we proposed a new evaluation framework which quantitatively evaluates image-to-image translation models, in terms of both content and style. Our framework is based on two CNNs which measure the style and content aspects separately. Using this framework, our proposed model, GANILLA, achieves the best overall performance compared to the current state-of-the-art models.

\section{Acknowledgements}

The numerical calculations reported in this paper were fully performed at TUBITAK ULAKBIM, High Performance and Grid Computing Center (TRUBA resources).

\section*{References}
\bibliography{refs}

\begin{thebibliography}{10}
\expandafter\ifx\csname url\endcsname\relax
  \def\url#1{\texttt{#1}}\fi
\expandafter\ifx\csname urlprefix\endcsname\relax\def\urlprefix{URL }\fi
\expandafter\ifx\csname href\endcsname\relax
  \def\href#1#2{#2} \def\path#1{#1}\fi

\bibitem{gatsy}
L.~A. Gatys, A.~S. Ecker, M.~Bethge, A neural algorithm of artistic style,
  arXiv preprint arXiv:1508.06576.

\bibitem{gatsy2}
L.~A. Gatys, A.~S. Ecker, M.~Bethge, Image style transfer using convolutional
  neural networks, {IEEE} Conference on Computer Vision and Pattern
  Recognition.

\bibitem{cyclegan}
J.-Y. Zhu, T.~Park, P.~Isola, A.~A. Efros, Unpaired image-to-image translation
  using cycle-consistent adversarial networks, {IEEE} International Conference
  on Computer Vision.

\bibitem{dualgan}
Z.~Yi, H.~Zhang, P.~Tan, M.~Gong, Dualgan: Unsupervised dual learning for
  image-to-image translation, {IEEE} International Conference on Computer
  Vision.

\bibitem{cartoongan}
Y.~Chen, Y.-K. Lai, Y.-J. Liu, Cartoongan: Generative adversarial networks for
  photo cartoonization, in: {IEEE} Conference on Computer Vision and Pattern
  Recognition, 2018.

\bibitem{onio1}
G.~Berger, R.~Memisevic, Incorporating long-range consistency in cnn-based
  texture generation, arXiv preprint arXiv:1606.01286.

\bibitem{onio2}
S.~Li, X.~Xu, L.~Nie, T.-S. Chua, Laplacian-steered neural style transfer, in:
  {ACM} International Conference on Multimedia, 2017, pp. 1716--1724.

\bibitem{onio3}
E.~Risser, P.~Wilmot, C.~Barnes, Stable and controllable neural texture
  synthesis and style transfer using histogram losses, arXiv preprint
  arXiv:1701.08893.

\bibitem{ofio1}
D.~Ulyanov, V.~Lebedev, A.~Vedaldi, V.~S. Lempitsky, Texture networks:
  Feed-forward synthesis of textures and stylized images., in: International
  Conference on Machine Learning, 2016, pp. 1349--1357.

\bibitem{ofio2}
J.~Johnson, A.~Alahi, L.~Fei-Fei, Perceptual losses for real-time style
  transfer and super-resolution, in: {IEEE} European Conference on Computer
  Vision, Springer, 2016, pp. 694--711.

\bibitem{ofio3}
C.~Li, M.~Wand, Precomputed real-time texture synthesis with markovian
  generative adversarial networks, in: {IEEE} European Conference on Computer
  Vision, Springer, 2016, pp. 702--716.

\bibitem{ofio1-4}
Y.~Li, C.~Fang, J.~Yang, Z.~Wang, X.~Lu, M.-H. Yang, Diversified texture
  synthesis with feed-forward networks, in: {IEEE} Conference on Computer
  Vision and Pattern Recognition, 2017.

\bibitem{stargan}
Y.~Choi, M.~Choi, M.~Kim, J.-W. Ha, S.~Kim, J.~Choo, Stargan: Unified
  generative adversarial networks for multi-domain image-to-image translation,
  in: {IEEE} Conference on Computer Vision and Pattern Recognition, 2018, pp.
  8789--8797.

\bibitem{levent}
L.~Karacan, Z.~Akata, A.~Erdem, E.~Erdem, Learning to generate images of
  outdoor scenes from attributes and semantic layouts (2016).
\newblock \href {http://arxiv.org/abs/1612.00215} {\path{arXiv:1612.00215}}.

\bibitem{pix2pix}
P.~Isola, J.-Y. Zhu, T.~Zhou, A.~A. Efros, Image-to-image translation with
  conditional adversarial networks, {IEEE} Conference on Computer Vision and
  Pattern Recognition.

\bibitem{lee2018diverse}
H.-Y. Lee, H.-Y. Tseng, J.-B. Huang, M.~Singh, M.-H. Yang, Diverse
  image-to-image translation via disentangled representations, in: {IEEE}
  European Conference on Computer Vision, 2018, pp. 35--51.

\bibitem{cho2019image}
W.~Cho, S.~Choi, D.~K. Park, I.~Shin, J.~Choo, Image-to-image translation via
  group-wise deep whitening-and-coloring transformation, in: {IEEE} Conference
  on Computer Vision and Pattern Recognition, 2019, pp. 10639--10647.

\bibitem{amodio2019travelgan}
M.~Amodio, S.~Krishnaswamy, Travelgan: Image-to-image translation by
  transformation vector learning, in: {IEEE} Conference on Computer Vision and
  Pattern Recognition, 2019, pp. 8983--8992.

\bibitem{tomei2019art2real}
M.~Tomei, M.~Cornia, L.~Baraldi, R.~Cucchiara, Art2real: Unfolding the reality
  of artworks via semantically-aware image-to-image translation, in: {IEEE}
  Conference on Computer Vision and Pattern Recognition, 2019, pp. 5849--5859.

\bibitem{liu2019few}
M.-Y. Liu, X.~Huang, A.~Mallya, T.~Karras, T.~Aila, J.~Lehtinen, J.~Kautz,
  Few-shot unsupervised image-to-image translation, in: {IEEE} International
  Conference on Computer Vision, 2019.

\bibitem{liu2017unsupervised}
M.-Y. Liu, T.~Breuel, J.~Kautz, Unsupervised image-to-image translation
  networks, in: Advances in Neural Information Processing Systems, 2017, pp.
  700--708.

\bibitem{huang2018multimodal}
X.~Huang, M.-Y. Liu, S.~Belongie, J.~Kautz, Multimodal unsupervised
  image-to-image translation, in: {IEEE} European Conference on Computer
  Vision, 2018, pp. 172--189.

\bibitem{sketchygan}
W.~Chen, J.~Hays, Sketchygan: Towards diverse and realistic sketch to image
  synthesis, in: {IEEE} Conference on Computer Vision and Pattern Recognition,
  2018.

\bibitem{alharbi2019latent}
Y.~Alharbi, N.~Smith, P.~Wonka, Latent filter scaling for multimodal
  unsupervised image-to-image translation, in: {IEEE} Conference on Computer
  Vision and Pattern Recognition, 2019, pp. 1458--1466.

\bibitem{liu2019sketchgan}
F.~Liu, X.~Deng, Y.-K. Lai, Y.-J. Liu, C.~Ma, H.~Wang, Sketchgan: Joint sketch
  completion and recognition with generative adversarial network, in: {IEEE}
  Conference on Computer Vision and Pattern Recognition, 2019, pp. 5830--5839.

\bibitem{ghosh2019isketchnfill}
A.~Ghosh, R.~Zhang, P.~K. Dokania, O.~Wang, A.~A. Efros, P.~H.~S. Torr,
  E.~Shechtman, Interactive sketch \& fill: Multiclass sketch-to-image
  translation, in: {IEEE} International Conference on Computer Vision, 2019.

\bibitem{pix2pixHD}
T.-C. Wang, M.-Y. Liu, J.-Y. Zhu, A.~Tao, J.~Kautz, B.~Catanzaro,
  High-resolution image synthesis and semantic manipulation with conditional
  gans, in: {IEEE} Conference on Computer Vision and Pattern Recognition, 2018,
  pp. 8798--8807.

\bibitem{draw}
S.~Hicsonmez, N.~Samet, F.~Sener, P.~Duygulu, Draw: Deep networks for
  recognizing styles of artists who illustrate children's books, in: ACM
  International Conference on Multimedia Retrieval, ACM, 2017, pp. 338--346.

\bibitem{gan-metrics}
A.~Borji, Pros and cons of gan evaluation measures, Computer Vision and Image
  Understanding 179 (2019) 41--65.

\bibitem{survey1}
Y.~Jing, Y.~Yang, Z.~Feng, J.~Ye, Y.~Yu, M.~Song, Neural style transfer: A
  review, arXiv preprint arXiv:1705.04058.

\bibitem{survey2}
H.~Huang, P.~S. Yu, C.~Wang, An introduction to image synthesis with generative
  adversarial nets, arXiv preprint arXiv:1803.04469.

\bibitem{gan}
I.~Goodfellow, J.~Pouget-Abadie, M.~Mirza, B.~Xu, D.~Warde-Farley, S.~Ozair,
  A.~Courville, Y.~Bengio, Generative adversarial nets, in: Advances in Neural
  Information Processing Systems, 2014, pp. 2672--2680.

\bibitem{gan2}
J.~Zhao, M.~Mathieu, Y.~LeCun, Energy-based generative adversarial network,
  arXiv preprint arXiv:1609.03126.

\bibitem{dcgan}
A.~Radford, L.~Metz, S.~Chintala,
  \href{http://arxiv.org/abs/1511.06434}{Unsupervised representation learning
  with deep convolutional generative adversarial networks}, CoRR
  abs/1511.06434.
\newline\urlprefix\url{http://arxiv.org/abs/1511.06434}

\bibitem{discogan}
T.~Kim, M.~Cha, H.~Kim, J.~K. Lee, J.~Kim, Learning to discover cross-domain
  relations with generative adversarial networks, CoRR abs/1703.05192.

\bibitem{resnet}
K.~He, X.~Zhang, S.~Ren, J.~Sun, Deep residual learning for image recognition,
  in: {IEEE} Conference on Computer Vision and Pattern Recognition, 2016, pp.
  770--778.

\bibitem{IN}
D.~Ulyanov, A.~Vedaldi, V.~S. Lempitsky, Instance normalization: The missing
  ingredient for fast stylization, CoRR abs/1607.08022.

\bibitem{Ledig_2017}
C.~Ledig, L.~Theis, F.~Huszar, J.~Caballero, A.~Cunningham, A.~Acosta,
  A.~Aitken, A.~Tejani, J.~Totz, Z.~Wang, et~al., Photo-realistic single image
  super-resolution using a generative adversarial network, in: {IEEE}
  Conference on Computer Vision and Pattern Recognition, IEEE, 2017.

\bibitem{li2016precomputed}
C.~Li, M.~Wand, Precomputed real-time texture synthesis with markovian
  generative adversarial networks, in: European Conference on Computer Vision,
  Springer, 2016, pp. 702--716.

\bibitem{goodfellow2014generative}
I.~Goodfellow, J.~Pouget-Abadie, M.~Mirza, B.~Xu, D.~Warde-Farley, S.~Ozair,
  A.~Courville, Y.~Bengio, Generative adversarial nets, in: Advances in Neural
  Information Processing Systems, 2014, pp. 2672--2680.

\bibitem{taigman2016unsupervised}
Y.~Taigman, A.~Polyak, L.~Wolf, Unsupervised cross-domain image generation,
  International Conference on Learning Representations.

\bibitem{pytorch}
A.~Paszke, S.~Gross, S.~Chintala, G.~Chanan, E.~Yang, Z.~DeVito, Z.~Lin,
  A.~Desmaison, L.~Antiga, A.~Lerer, Automatic differentiation in pytorch.

\bibitem{SUN}
J.~Xiao, J.~Hays, K.~Ehinger, A.~Oliva, A.~Torralba, Sun database: Large-scale
  scene recognition from abbey to zoo, in: {IEEE} Conference on Computer Vision
  and Pattern Recognition, 2010, pp. 3485--3492.

\end{thebibliography}

\end{document}